\documentclass{egpubl}
\usepackage{3dor11}

\usepackage{amsmath,amssymb}

 \WsPaper         
 \electronicVersion 

\ifpdf \usepackage[pdftex]{graphicx} \pdfcompresslevel=9
\else \usepackage[dvips]{graphicx} \fi

\PrintedOrElectronic

\usepackage{t1enc,dfadobe}

\usepackage{egweblnk}
\usepackage{cite}

\title[SHREC'11: robust feature detection/description]%
      {SHREC 2011: robust feature detection and description benchmark}

\author[Boyer {\em et al.}]
       {
       E. Boyer$^{1}$, 
       A.~M. Bronstein\thanks{Organizer of the SHREC track. All organizers and participants are listed in alphabetical order. For any information about the benchmark, contact michael.bronstein@usi.ch. Authors are listed alphabetically. }$^{2}$,
        M.~M. Bronstein$^{\dagger 3}$,
B. Bustos$^4$,
T. Darom$^{5}$, 
R. Horaud$^{1}$,
I. Hotz$^{6}$, 
Y. Keller$^{5}$, 
J. Keustermans$^{7}$,\and
A. Kovnatsky$^{\dagger 8}$,
R. Litman$^{\dagger 2}$,
J. Reininghaus$^{6}$, 
I. Sipiran$^4$,
D. Smeets$^{7}$, 
P. Suetens$^{7}$, 
D. Vandermeulen$^{7}$,\and
A. Zaharescu$^{\dagger 9}$, 
V. Zobel$^{6}$
        \\
$^1$INRIA Grenoble Rh{\^o}ne-Alpes, France\\
$^2$Department of Electrical Engineering, Tel Aviv University, Israel\\
$^3$Institute of Computational Science, Faculty of Informatics, Universit{\`a} della Svizzera Italiana, Lugano, Switzerland\\
$^4$Department of Computer Science, University of Chile\\
$^5$School of Engineering, Bar-Ilan University, Ramat-Gan, Israel\\
$^6$Zuse Institut Berlin, Germany\\
$^7$Department of Electrical Engineering, K.U. Leuven, Belgium \\
$^8$Department of Mathematics, Technion -- Israel Institute of Technology, Haifa, Israel\\
$^9$Aimetis Corp., Waterloo, Canada\\
       }

\begin{document}

\maketitle

\begin{abstract}
Feature-based approaches have recently become very popular in computer vision and image analysis applications, and are becoming a promising direction in shape retrieval.
SHREC'11 robust feature detection and description benchmark simulates the feature detection and description stages of feature-based shape retrieval algorithms. The benchmark tests the performance of shape feature detectors and descriptors under a wide variety of transformations. The benchmark allows evaluating how algorithms cope with certain classes of transformations and strength of the transformations that can be dealt with.
The present paper is a report of the SHREC'11 robust feature detection and description benchmark results.


\end{abstract}

\section{Introduction}

Feature-based approaches have recently become very popular in computer vision and image analysis applications, notably due to the works of
Lowe \cite{lowe2004distinctive}, Sivic and Zisserman \cite{sivic2003video}, and Mikolajczyk and Schmid \cite{mikolajczyk2005performance}. In these approaches, an image is described as a collection of local features (``visual words'')
from a given vocabulary, resulting in a representation referred to as a \emph{bag of features}.
The bag of features paradigm relies heavily on the choice of the local feature descriptor that is used to create the visual words.
A common evaluation strategy of image feature detection and description algorithms is the stability of the detected features and their invariance to different transformations applied to an image.
In shape analysis, feature-based approaches have been introduced more recently and are gaining popularity in shape retrieval applications. 

SHREC'11 invariant feature detection and description benchmark simulates the feature detection and description stages of feature-based shape retrieval algorithms. The benchmark tests the performance of shape feature detectors and descriptors under a wide variety of different transformations. The benchmark allows evaluating how algorithms cope with certain classes of transformations and what is the strength of the transformations that can be dealt with.

This report presents a long version of the paper \cite{SHREC11f}.

\section{Data}

The dataset used in this benchmark was from the TOSCA shapes \cite{bronstein2008numerical}, available in the public domain.
The shapes were represented as triangular meshes with approximately 10,000--50,000 vertices.

\begin{figure*}[tpb]
  \centering \includegraphics*[width=1\linewidth]{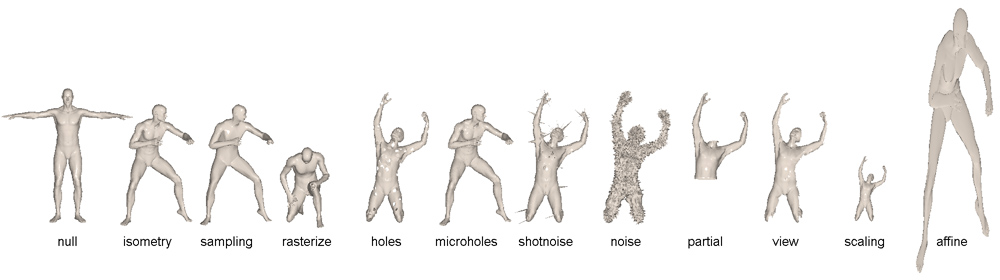}
   \caption{\label{fig:shapes1} Transformations of the human shape used in the tests (shown in strength 5, left to right): null, isometry, sampling, rasterize, holes, micro holes, shot noise, noise, partial, view, scaling, affine.}
\end{figure*}

The dataset includes ones shape class (human) with simulated transformations. Compared to the SHREC 2010 benchmark, there are additional transformation classes and the transformations themselves are more challenging. 
For each null shape, transformations were split into 11 classes shown in Figure~\ref{fig:shapes1}:
{\em isometry} (non-rigid triangulation- and distance-preserving almost inelastic deformations),
{\em topology} (welding of shape vertices resulting in different triangulation),
{\em rasterization} (simulating non-pointwise topological artifacts due to occlusions in 3D geometry acquisition), {\em view} (simulating missing parts due to 3D acquisition artifacts), {\em partial} (missing parts), 
{\em micro holes} and big {\em holes},
global uniform {\em scaling}, global {\em affine} transformations, additive Gaussian {\em noise}, {\em shot noise},
down-{\em sampling} (less than 20\% of the original points).

In each class, the transformation appeared in five different versions numbered 1--5. In all shape categories except scale and isometry,
the version number corresponded to the transformation strength levels: the higher the number, the stronger the transformation (e.g., in noise transformation, the noise variance was proportional to the strength number). For the isometry class, the numbers do not reflect the strength of the transformation.
The total number of transformations was 55.
The dataset is available at http://tosca.cs.technion.ac.il/book/shrec\_feat.html.

\section{Evaluation methodology}

The evaluation was performed separately for feature detection and feature description algorithms. Feature detectors were further divided into point and region; feature descriptors were divided into point, region, and dense. 
The participants were asked to provide, for each shape $Y$ in the dataset, (i) a set of detected feature points $\mathcal{F}(Y) = \{y_k\in Y\}_{k}$ or regions $\mathcal{F}(Y) = \{Y_l\subset Y\}_{l}$;
(ii) optionally, for each detected point $y_k$, a descriptor vector $\{\mathbf{f}(y_k)\}_{k=1}^{|\mathcal{F}(Y)|}$; or alternatively, for each detected region $Y_l$, a descriptor vector $\{\mathbf{f}(Y_l)\}_{l=1}^{|\mathcal{F}(Y)|}$. 
For dense descriptors, participants provided $\{\mathbf{f}(y_k)\}_{k=1}^{|Y|}$.
The performance was measured by comparing features and feature descriptors computed for transformed shapes and the corresponding null shapes.

\subsection{Feature detection}
The quality of the feature detection was measured using the \emph{repeatability} criterion.
Assuming for each transformed shape $Y$ in the dataset the groundtruth dense correspondence to the null shape $X$ to be given in the form of pairs of points
$\mathcal{C}_0(X,Y) = \{(x'_k,y_k) \}_{k=1}^{|Y|}$, a feature point $y_k \in \mathcal{F}(Y)$ is said
to be \emph{repeatable} if a geodesic ball of radius $\rho$ around the corresponding point $x'_k : (x'_k,y_k) \in \mathcal{C}_0(X,Y)$ contains
a detected feature point $x_j \in \mathcal{F}(X)$.\footnote{Features without groundtruth correspondence (e.g. in regions in the null shape corresponding to holes in the transformed shape) are ignored.}
Repeatable features are  
\begin{eqnarray*}
\mathcal{F}_{\rho}(Y) &=& \{y_k \in \mathcal{F}(Y) : \mathcal{F}(X) \cap B_\rho(x'_k) \neq \emptyset, \,\,\, \\
&&(x'_k,y_k) \in \mathcal{C}_0(X,Y) \},
\end{eqnarray*}
 where $B_\rho(x'_k) = \{ x\in X : d_X(x,x'_k) \leq \rho \}$ and $d_X$ denotes the geodesic distance function in $X$.

Similarly, for region detectors, a region $Y_l \in \mathcal{F}(Y)$ is repeatable if the corresponding region $X'_l \subset X$ has overlap larger than $\rho$, 
\begin{eqnarray*}
\mathcal{F}_{\rho}(Y) &=& \{Y_l \in \mathcal{F}(Y) : |X'_l \cap X_l| / | X'_l \cup X_l | \geq \rho \}. 
\end{eqnarray*}

The {\em repeatability} of a feature detector is defined as the percentage $|\mathcal{F}_{\rho}(Y)|/|\mathcal{F}(Y)|$ of features that are repeatable, the definition being dependent of whether a point or region descriptor is used. 
%
%
%

\subsection{Feature description}

Let $\{\mathbf{f}_k\}_{k=1}^{|\mathcal{F}(Y)|}$, $\{\mathbf{g}_j)\}_{j=1}^{|\mathcal{F}(X)|}$ denote descriptors computed on feature points 
$\mathcal{F}(X)$ and $\mathcal{F}(Y)$, respectively. 
For point descriptors, we consider as the point corresponding to $y_k$ the closest point $x_j \in \mathcal{F}(X)$ to $x'_k$, where $(x'_k,y_k) \in \mathcal{C}_0(X,Y)$, such that $r_{kj} = d_X(x_j,x'_k) < \rho$ for some $\rho$. 
%

Descriptor quality was evaluated using the normalized $L_2$ distance between descriptors at corresponding points, 
$$d_{kj} =  \frac{\| \mathbf{f}_k - \mathbf{g}_j \|_2}{ \frac{1}{|\mathcal{F}(X)|^2-|\mathcal{F}(X)|}\sum_{k,j\neq k} \| \mathbf{f}_k - \mathbf{g}_j \|_2}.$$  
%

%

In addition, an evaluation using the ROC was performed as follows. 
The corresponding feature points $x_k, y_j$ 
are considered {\em true positives} if $d_{kj} \leq \tau$, for some threshold $\tau$. 
The {\em true positive rate} is defined as 
$TPR = |\{  d_{kj} \leq \tau \}|/|\{ r_{kj} \leq \rho \}|$; 
the {\em false positive rate} is defined as 
$FPR = |\{  d_{kj} \leq \tau \}|/|\{ r_{kj} > \rho \}|$. 
By varying the threshold $\tau$, a set of pairs $(FPR, TPR)$ referred to as the {\em receiver operation characteristic} (ROC) curve is obtained. For a fixed FPR, the higher the TPR, the better.

For a dense descriptor, 
%
the quality is measured as the average normalized $L_2$ distance between the descriptor vectors in corresponding points,
$$\frac{1}{|\mathcal{F}(X)|}\sum_{k=1}^{|\mathcal{F}(X)|} d_{kj}. $$ 
%

\section{Feature detection methods}

\subsection{Point features}

\textbf{Harris 3D} (Sipiran and Bustos \cite{sipiran}). 
The algorithm proposes an extension for meshes of the Harris corner detection method \cite{harris}. The algorithm suggests to determine a neighborhood (rings or adaptive)
around a vertex. Next, this neighborhood is used to fit a quadratic patch which is considered as an image. After applying a gaussian smoothing, derivatives are calculated which are used 
to calculate the Harris response for each vertex.
In this benchmark, three different configurations were used: adaptive neighborhoods with $\delta=0.01$, 1-ring neighborhoods , and 2-ring neighborhoods.
For details, see \cite{sipiran}.


\textbf{Mesh-DoG} (Zaharescu {\em et al.} \cite{Zaharescu:2009cvpr} ). 
The method considers the general setting of 2-D manifolds $\mathcal{M}$ embedded in $\mathbb{R}^3$ endowed in with a scalar function $f:\mathcal{M} \rightarrow \mathbb{R}$, such as colour or curvature. This represents a generalization of 2-D images, that can be viewed as a uniformly sampled square grid with vertices of valence 4. Operators, such as the gradient and the convolution are defined in this context.
A scale-space representation of the scalar function $f$ is build using iterative convolutions with a Guassian kernel.  Feature detection consists of two steps. Firstly, the extrema of the function's Laplacian (approximated by taking the difference between adjacent scales -  Difference of Gaussian) are found across scales, followed by non-maximum suppression using a 1-ring neighbourhood both spatially and across adjacent scales. Secondly, the detected extrema are thresholded (400 points).
Mean  and Gaussian curvature computed using \cite{Meyer:2002} were the scalar functions used for current tests. 
For exact details and settings, see \cite{Zaharescu:2009cvpr}.

\textbf{Mesh SIFT} (Smeets {\em et al.} \cite{Maes2010}). 
The Mesh SIFT detector detects scale space extrema as local feature locations. 
First, a scale space is constructed containing smoothed versions of the input mesh, which are obtained by subsequent convolutions of the mesh with a binomial filter. 
Next, for the detection of salient points in the scale space, the mean curvature $H$ (Mesh SIFT-H) and the principal coordinates in curvature space $KK$ (Mesh SIFT-KK), which are minimal and maximal curvature, are computed for each vertex and at each scale in the scale space ($H_i$ and $KK_i$). Note that the mesh is smoothed and not the function on the mesh ($H$ or $KK$).
Scale space extrema in scale spaces of differences between subsequent scales ($dH_i = H_{i+1} - H_i$ for Mesh SIFT-H and  $ dKK_i = KK_{i+1} - KK_i$ for Mesh SIFT-KK) are finally selected as local feature locations.

\textbf{Mesh-Scale DoG} (Darom and Keller~\cite{Darom11})
We follow the work of Zaharescu \emph{et al.}~\cite{Zaharescu:2009cvpr} that presented a Difference of Gaussians based feature points detector for mesh objects.
We propose to define a Gaussian filter on the mesh geometry, and
compute a set of filtered meshes. Consecutive octaves are subtracted to compute
the DoG function, and define the local maxima (both in location and scale) as
our feature points at that point and scale. In order to make the detected features scale invariant, we suggest to set the support for each feature point to the width of the filter at that scale.
For details, see \cite{Darom11}.

\subsection{Region detectors}

\textbf{Shape MSER} (Litman {\em et al.} \cite{litman2010diffusion}). 
The algorithm finds maximally stable components in 3D shapes, similarly to the popular MSER method for feature analysis in images \cite{matas2004robust}. The shape is represented as a component tree based on vertex- or edge-wise weighting function (VW and EW, respectively).
In this benchmark, three different weights were used: edge weighting by inverse of commute time kernel (EW 1/CT) and inverse heat kernel (EW 1/HKS), and vertex weighting by heat kernel diagonal (VW HKS). 
For details, see \cite{litman2010diffusion}.

\section{Feature description methods}

\subsection{Point descriptors}

\textbf{Mesh-HoG} (Zaharescu {\em et al.} \cite{Zaharescu:2009cvpr} ). 
For a given interest point, the descriptor is computed using a geodesic support region, proportional to $3\%$ of the total surface area.  For each vertex in the neighbourhood, the 3-D gradient information is computed using $f$ at the detected scale. As a first step, a local coordinate system is chosen, in order to make the descriptor rotation invariant. Then, a histogram of gradient is computed, both spatially, at a coarse level, in order to maintain a certain high-level spatial ordering, and using orientations, at a finer level. Since the gradient vectors are 3 dimensional, the histograms are computed in 3D. The histograms are concatenated and normalized. A $96$ dimensional descriptor is obtained. The gradient of the participating neighbouring vertices is computed at the scale of the detected interest point. 
For exact details and settings, see \cite{Zaharescu:2009cvpr}.


\textbf{Scale Invariant Spin Image} (Darom and Keller~\cite{Darom11})
The Spin Image local descriptor was presented by Johnson and
Hebert~\cite{JohnsonH99}, and has gained popularity due to its robustness and
simplicity. Utilizing the local scale estimated by the Mesh-Scale DoG detector, we
propose to derive a Scale Invariant Spin Image mesh descriptor, where we
compute the Spin Image descriptor over the local scale estimated at the interest point. This improves
feature point matching, in particular when the meshes are related significant
partial matching.
For details, see \cite{Darom11}.

\textbf{Local Depth SIFT} (Darom and Keller~\cite{Darom11})
The SIFT algorithm, presented by D.~Lowe~\cite{lowe2004distinctive} is a state-of-the-art
approach to computing scale and rotation invariant local features in images.
The SIFT descriptor is based on computing a local radial-angular histogram of the pixel value derivatives. Inspired by Lowe's seminal work, we propose to compute a new local feature for 3D meshes we denote \emph{Local Depth SIFT }(LD-SIFT).
Given an interest point we estimate its tangent plane, and compute the distance from each point on the
surface to that plane to create a depth map, and set the viewport size to match the feature scale, as detected by the Mesh-Scale DoG detector. This makes our construction scale invariant. We compute the PCA of the the points surrounding the interest point, and use their dominant direction as the \emph{local dominant angle}, 
and rotate the depth map to a canonical angle based on the dominant angle. 
This makes the LD-SIFT rotation invariant. We compute a SIFT feature descriptor on the resulting depth map to create the Local Depth SIFT feature descriptor.
For details, see \cite{Darom11}.

%

\subsection{Dense descriptors}

\textbf{Generalized HKS} (Zobel {\em et al.} \cite{Zobel2011}). 
The Generalized HKS is a generalization of the HKS \cite{sun09hks} to 1-forms (where a 1-form can be regarded as vector field). It is derived from the heat kernel for 1-forms in a similar way as the HKS is derived from the heat kernel for functions. This yields a symmetric tensor field of second order with a time parameter $t$. For easier comparability we consider scalar tensor invariants. For details see \cite{Zobel2011} or \cite{Zobel2010}.


\section{Results}

\subsection{Point feature detectors.}

\begin{figure*}[t]
    \includegraphics[width=\linewidth]{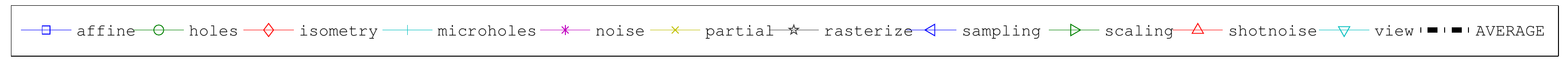} \\
     \begin{tabular}{cc}
     	 \includegraphics[width=\columnwidth]{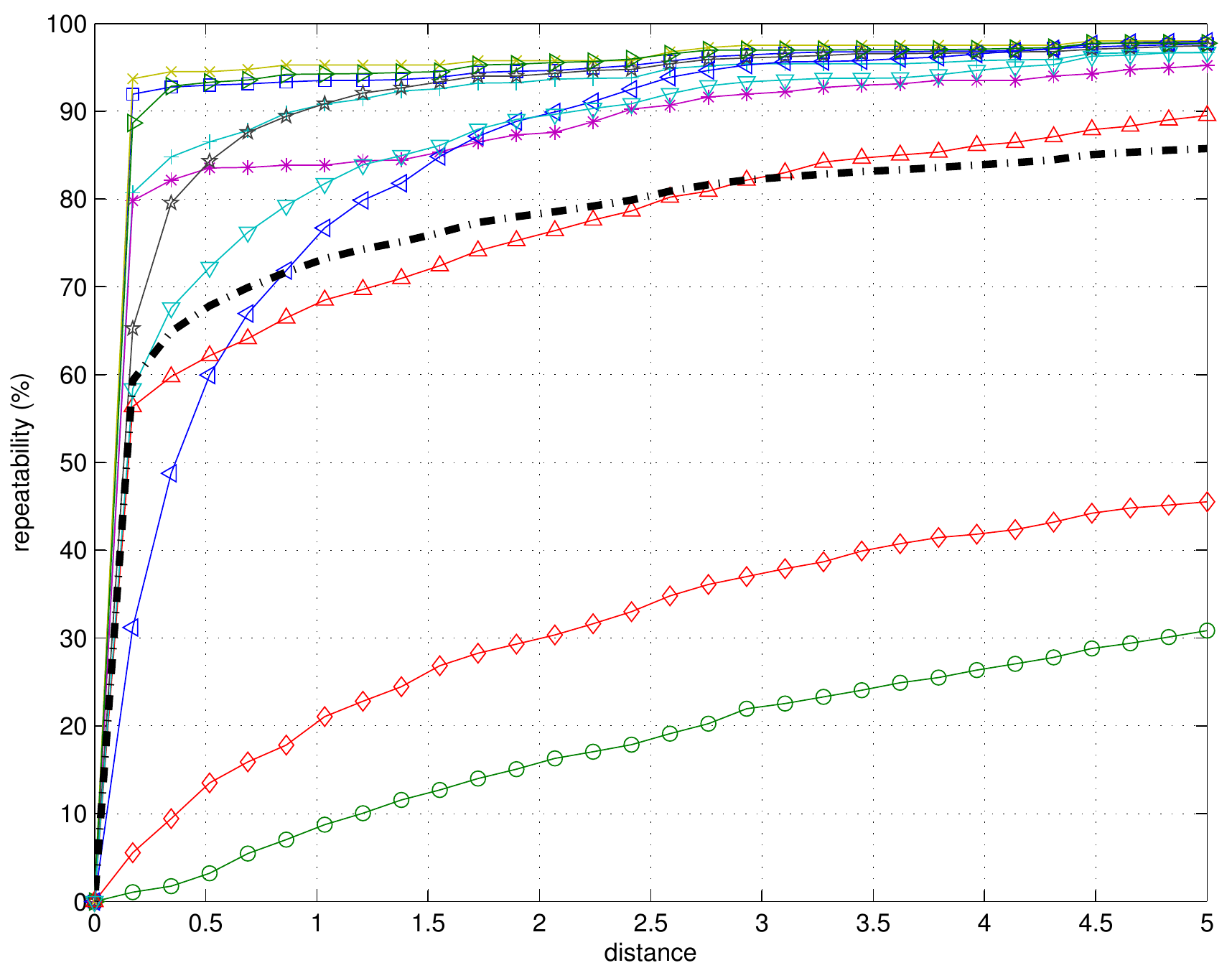} &
         \includegraphics[width=\columnwidth]{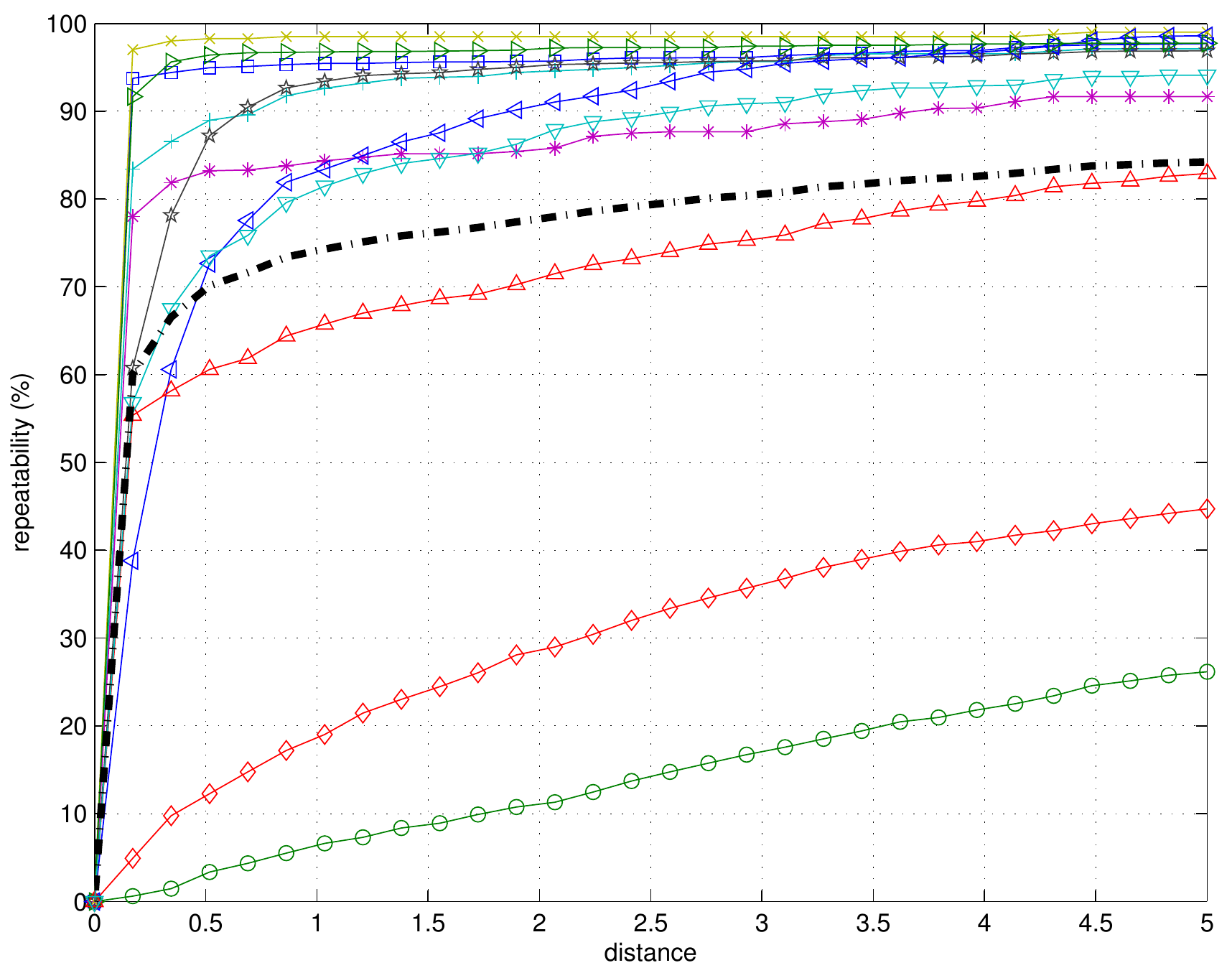} \\
 	{\small Mesh DoG (mean)} & {\small Mesh DoG (Gaussian)}\\   
	 \includegraphics[width=\columnwidth]{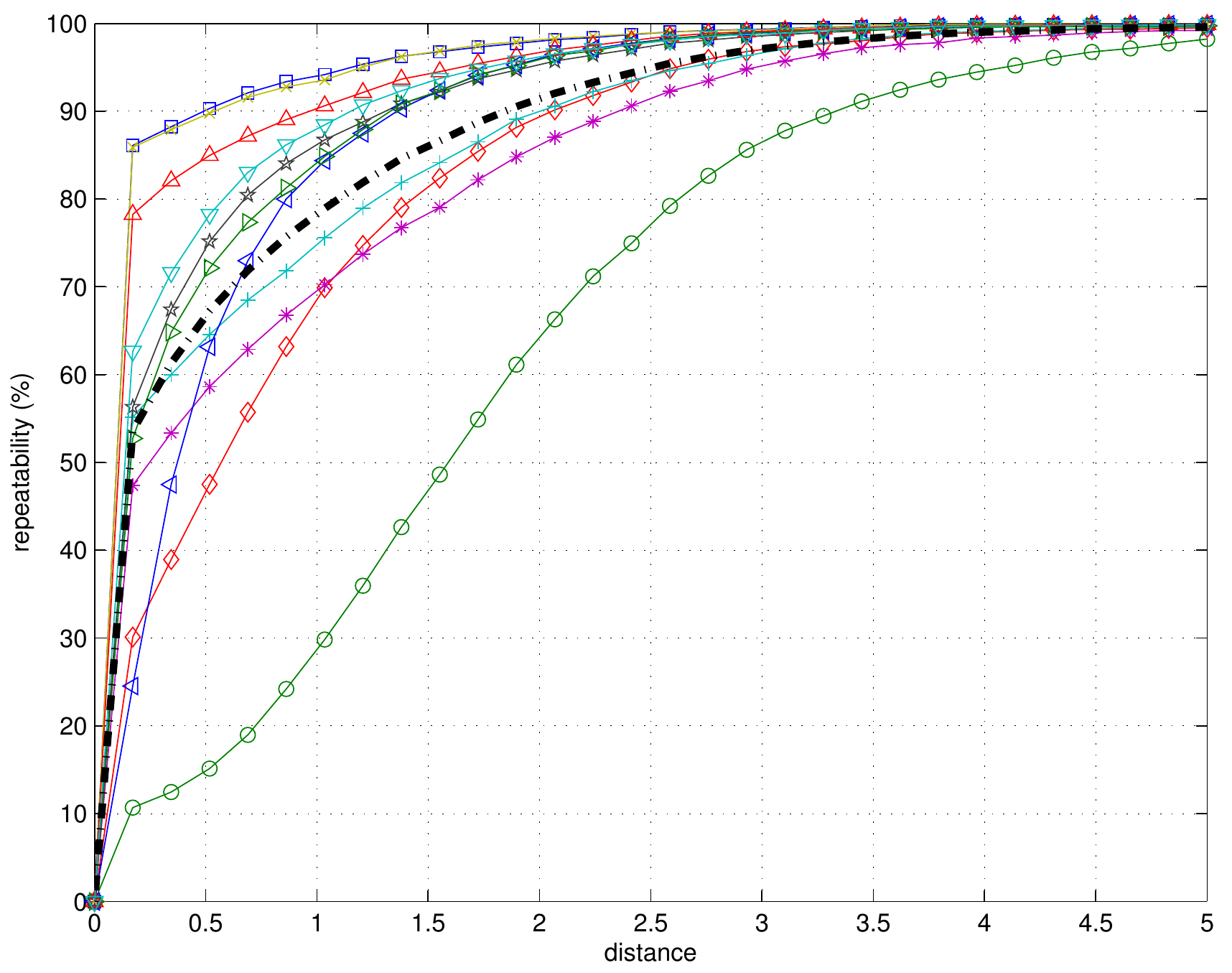} &
	 \includegraphics[width=\columnwidth]{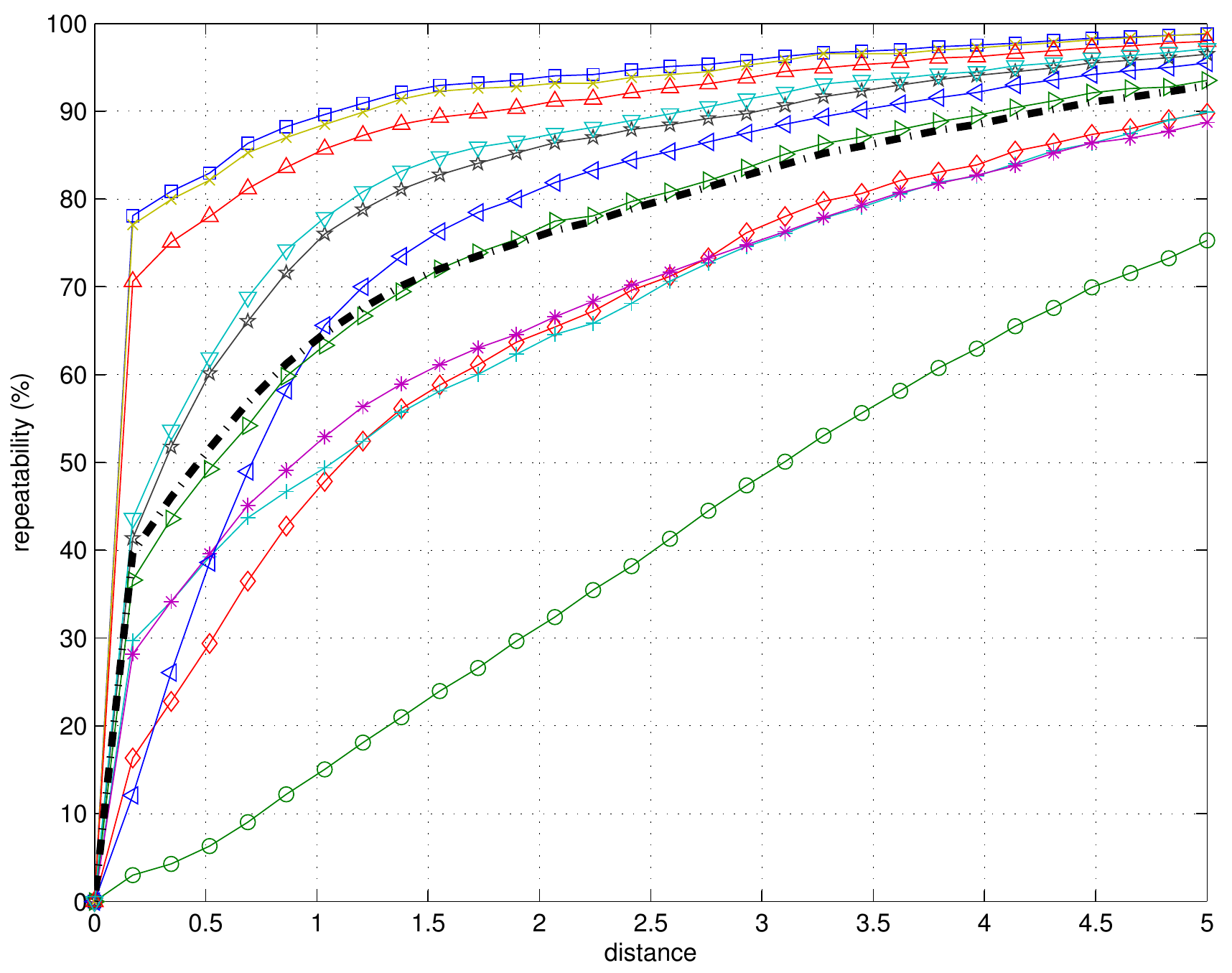} \\	 
 	{\small Mesh-Scale DoG (1)} & {\small IMesh-Scale DoG (2)}\\         
	 \includegraphics[width=\columnwidth]{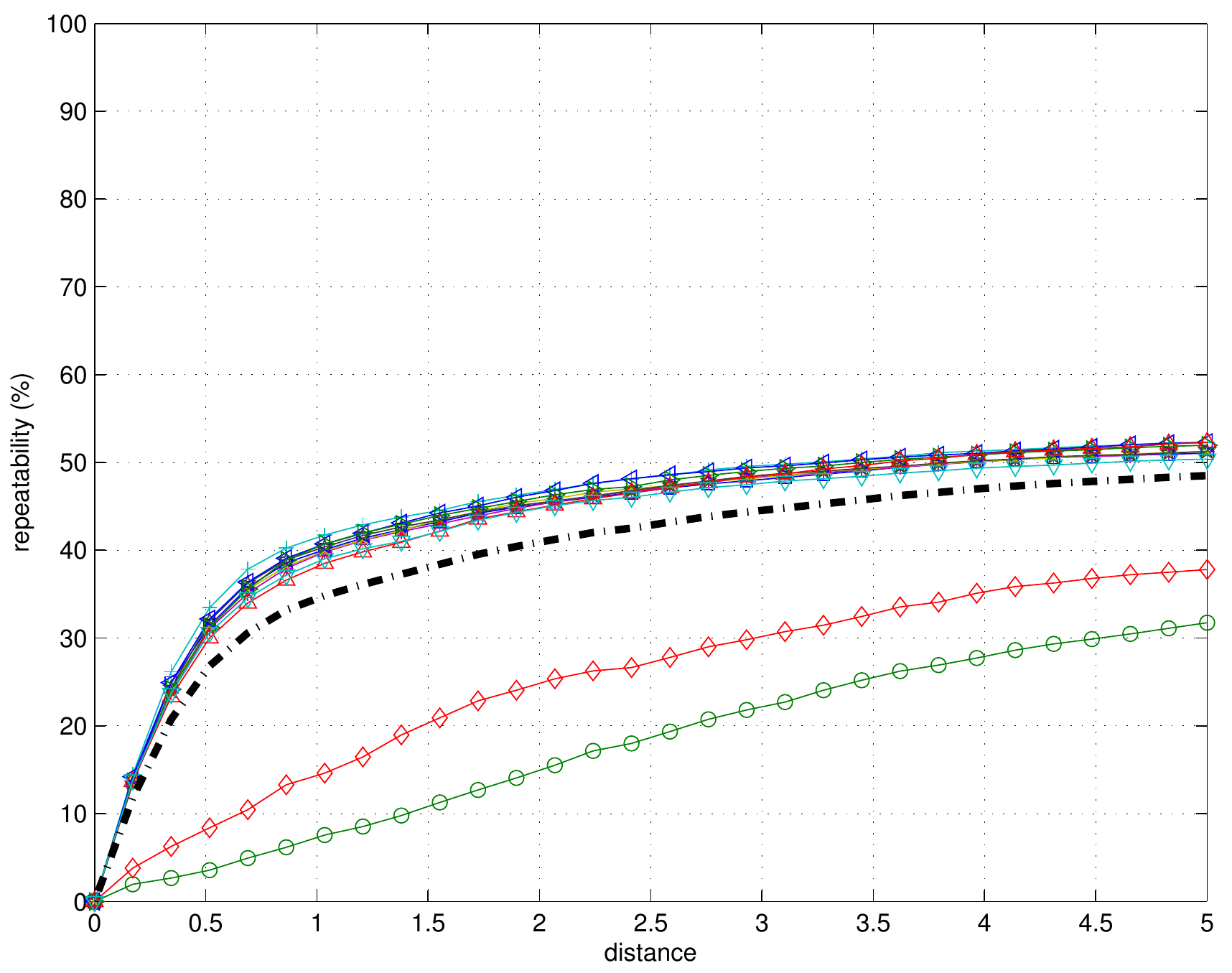} &
	 \includegraphics[width=\columnwidth]{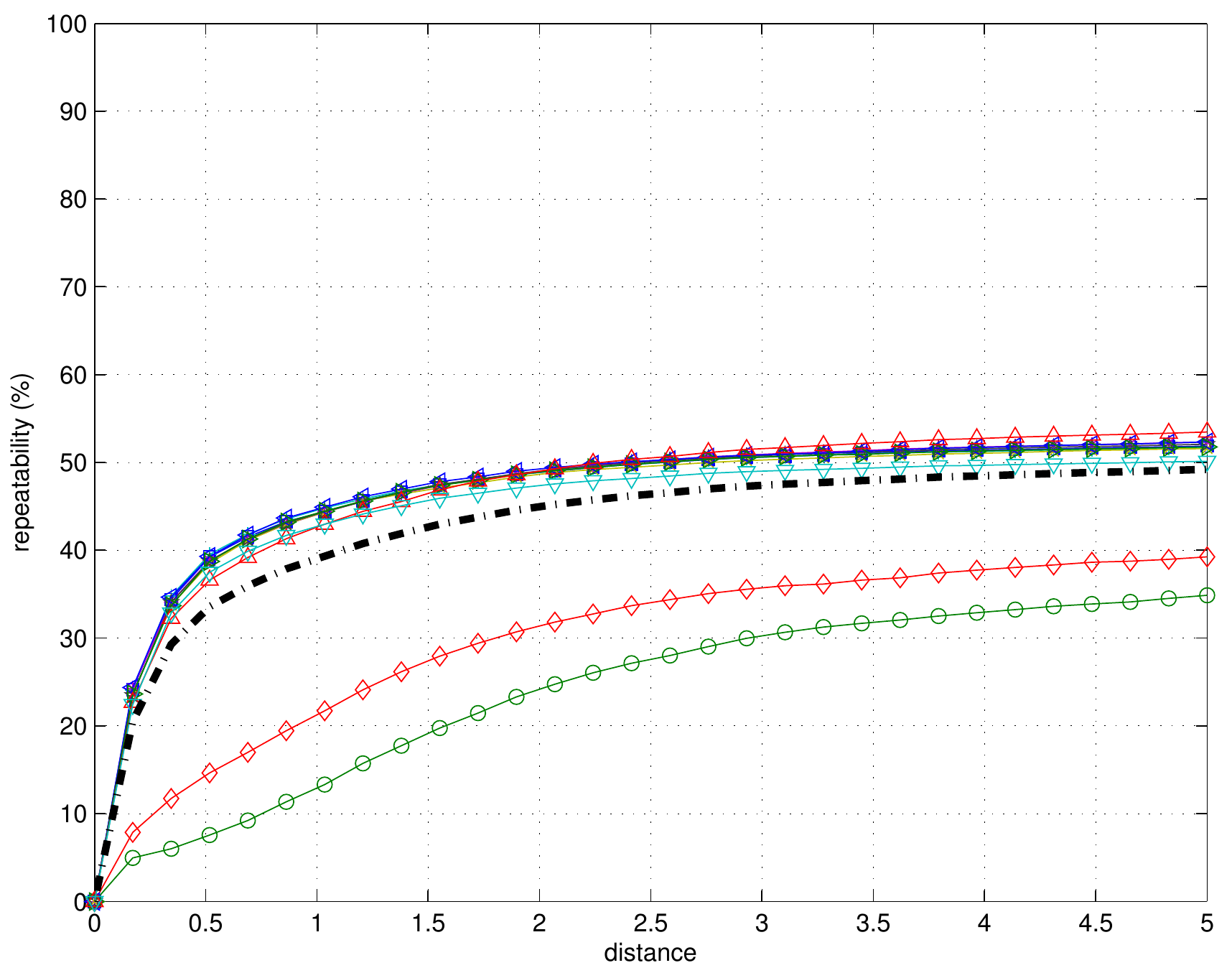} \\	 
 	{\small Mesh SIFT (H)} & {\small Mesh SIFT (KK)}\\         
	
    \end{tabular}
\caption{Repeatability ($\%$) vs distance of point feature detectors broken down according to different transformation classes.  }
\label{fig_repeatability_point}
\end{figure*}

\begin{figure*}[t]
    \includegraphics[width=\linewidth]{Legend_H-eps-converted-to.pdf} \\
     \begin{tabular}{cc}
        \includegraphics[width=\columnwidth]{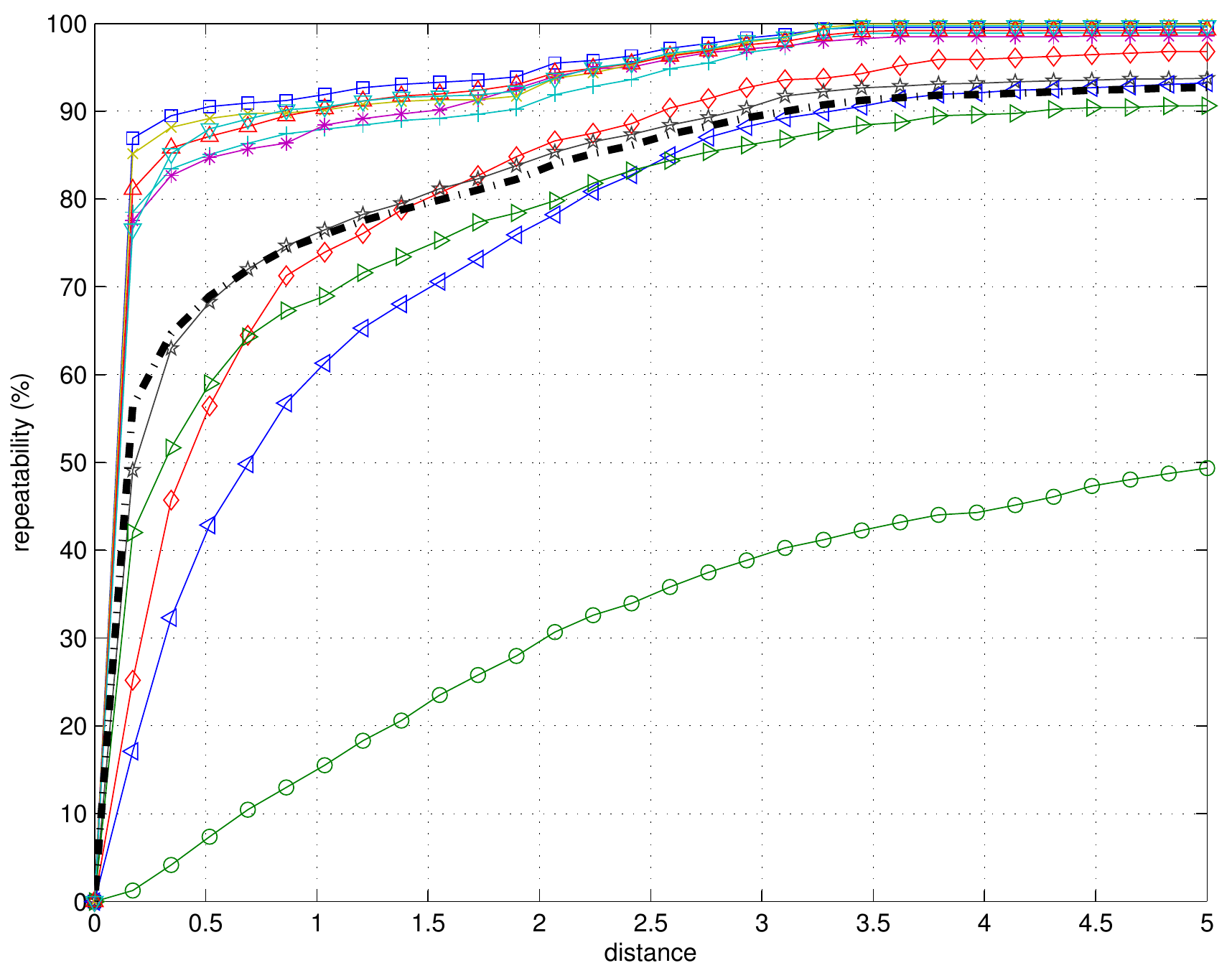} &
        \includegraphics[width=\columnwidth]{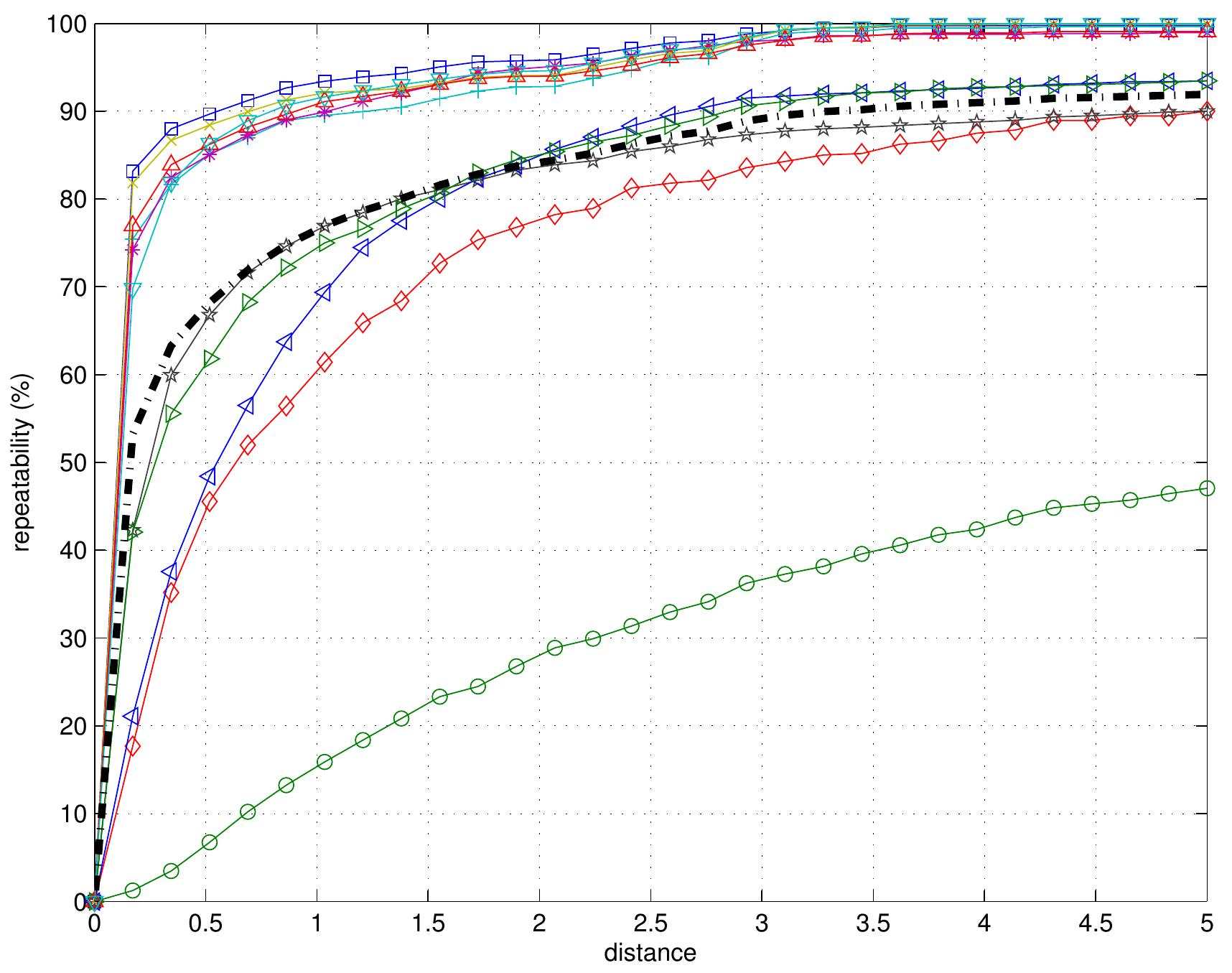} \\
        {\small 3D Harris (ring 1)} & {\small 3D Harris (ring 2)}\\                
        \multicolumn{2}{c}{\includegraphics[width=\columnwidth]{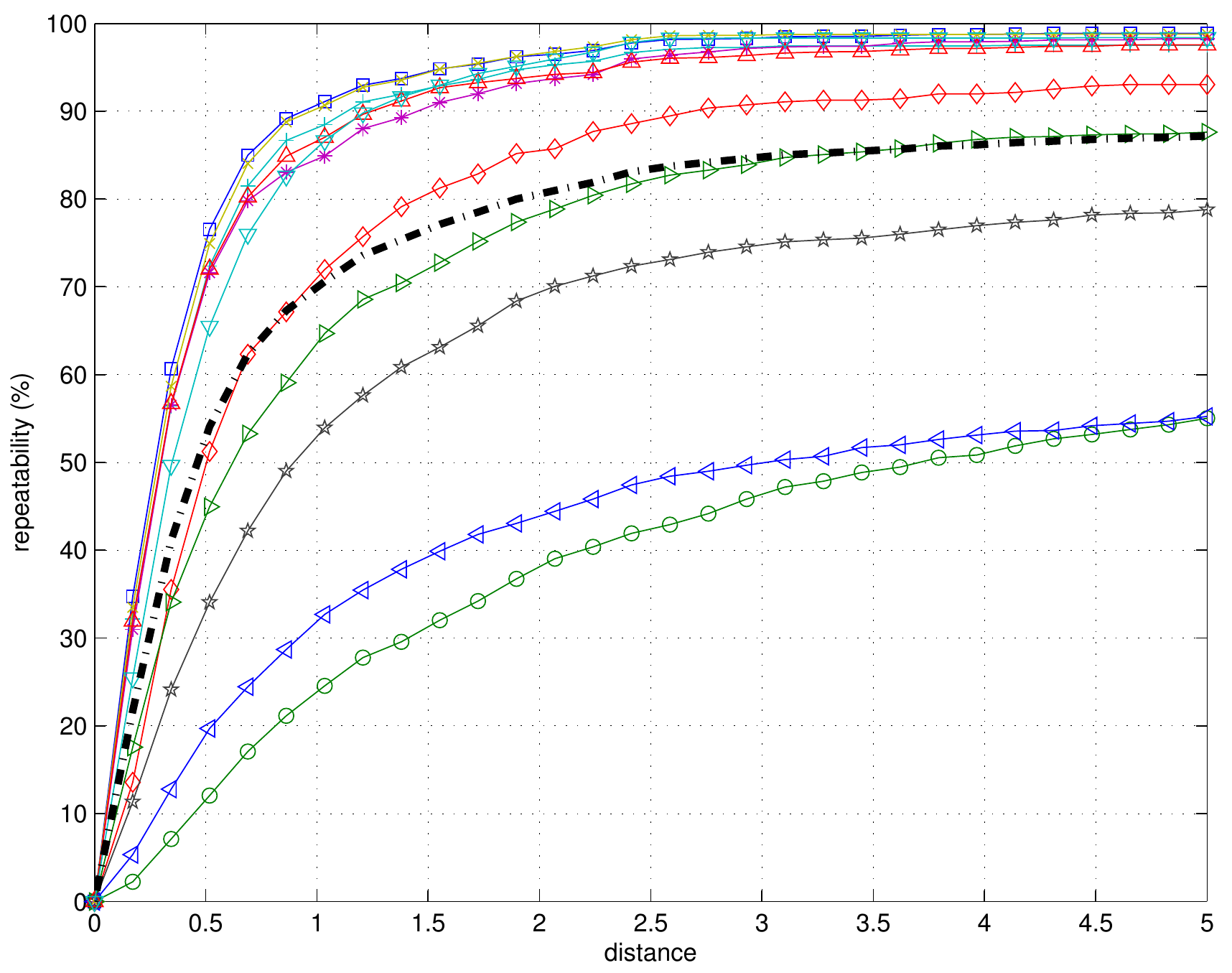}} \\
 	\multicolumn{2}{c}{\small 3D Harris (adaptive)} \\       
    \end{tabular}
\caption{Repeatability ($\%$) vs distance of point feature detectors broken down according to different transformation classes. }
\label{fig_repeatability_point1}
\end{figure*}

Tables~\ref{tab:Zaharescu_corr-det1-feat1-pt400}--\ref{tab:sipiran_ivan_Harris3DAdaptive} show the repeatability of different point descriptors at fixed radius $\rho=5$ (approximately $1\%$ of the shape diameter), broken down according to transformation classes and strengths. Higher repeatability scores are indication of better performance. 
Figures~\ref{fig_repeatability_point}--\ref{fig_repeatability_point1} show the repeatability of point descriptors as function of geodesic distance 
varying from $0$ to $5$. 

\begin{table}
\centering
\begin{tabular}{lccccc}
& \multicolumn{5}{c}{\small\textbf{ Strength}} \\
\cline{2-6}
{\small\textbf{ Transform.}} & {\small\textbf{  1}} & {\small\textbf{ $\leq$2}} & {\small\textbf{ $\leq$3}} & {\small\textbf{ $\leq$4}} & {\small\textbf{ $\leq$5}}\\
\hline
{\small\em Isometry} & {\small 97.75} & {\small 98.13} & {\small 97.92} & {\small 97.94} & {\small 97.70} \\
{\small\em Rasterization} & {\small 35.50} & {\small 37.75} & {\small 36.17} & {\small 34.19} & {\small 30.85} \\
{\small\em Sampling} & {\small 73.50} & {\small 59.50} & {\small 50.67} & {\small 44.72} & {\small 42.06} \\
{\small\em Holes} & {\small 96.50} & {\small 96.88} & {\small 96.83} & {\small 96.88} & {\small 96.65} \\
{\small\em Micro holes} & {\small 96.50} & {\small 95.75} & {\small 95.50} & {\small 95.38} & {\small 95.20} \\
{\small\em Scaling} & {\small 98.00} & {\small 98.00} & {\small 98.00} & {\small 98.00} & {\small 98.00} \\
{\small\em Affine} & {\small 98.25} & {\small 98.75} & {\small 98.50} & {\small 98.13} & {\small 97.45} \\
{\small\em Noise} & {\small 99.25} & {\small 99.13} & {\small 98.50} & {\small 98.25} & {\small 97.95} \\
{\small\em Shot Noise} & {\small 98.25} & {\small 98.00} & {\small 98.00} & {\small 97.87} & {\small 97.75} \\
{\small\em Partial} & {\small 98.25} & {\small 97.25} & {\small 97.17} & {\small 90.06} & {\small 89.50} \\
{\small\em View} & {\small 95.50} & {\small 96.38} & {\small 96.33} & {\small 97.00} & {\small 96.70} \\
\hline
{\small{\textbf{Average}}} & {\small 89.75} & {\small 88.68} & {\small 87.60} & {\small 86.22} & {\small 85.44} \\
\hline
\end{tabular}
\caption{\small Repeatability (in $\%$) at $\rho=5$ of Mesh DoG (mean) feature detection algorithm. Average number of detected points: 392.\label{tab:Zaharescu_corr-det1-feat1-pt400}}
\end{table}
\begin{table}
\centering
\begin{tabular}{lccccc}
& \multicolumn{5}{c}{\small\textbf{ Strength}} \\
\cline{2-6}
{\small\textbf{ Transform.}} & {\small\textbf{  1}} & {\small\textbf{ $\leq$2}} & {\small\textbf{ $\leq$3}} & {\small\textbf{ $\leq$4}} & {\small\textbf{ $\leq$5}}\\
\hline
{\small\em Isometry} & {\small 99.00} & {\small 99.38} & {\small 98.58} & {\small 98.69} & {\small 97.70} \\
{\small\em Rasterization} & {\small 29.50} & {\small 30.66} & {\small 31.03} & {\small 29.08} & {\small 26.17} \\
{\small\em Sampling} & {\small 76.00} & {\small 61.00} & {\small 50.50} & {\small 43.68} & {\small 39.77} \\
{\small\em Holes} & {\small 98.25} & {\small 98.25} & {\small 98.08} & {\small 97.94} & {\small 97.10} \\
{\small\em Micro holes} & {\small 94.00} & {\small 92.88} & {\small 92.42} & {\small 92.19} & {\small 91.65} \\
{\small\em Scaling} & {\small 99.00} & {\small 99.00} & {\small 99.00} & {\small 99.00} & {\small 99.00} \\
{\small\em Affine} & {\small 98.00} & {\small 98.25} & {\small 98.00} & {\small 97.69} & {\small 96.90} \\
{\small\em Noise} & {\small 99.75} & {\small 99.50} & {\small 99.00} & {\small 98.75} & {\small 98.60} \\
{\small\em Shot Noise} & {\small 98.25} & {\small 98.13} & {\small 98.00} & {\small 97.88} & {\small 97.75} \\
{\small\em Partial} & {\small 98.75} & {\small 96.38} & {\small 95.33} & {\small 85.31} & {\small 82.90} \\
{\small\em View} & {\small 92.50} & {\small 92.75} & {\small 92.08} & {\small 92.94} & {\small 94.10} \\
\hline
{\small{\textbf{Average}}} & {\small 89.36} & {\small 87.83} & {\small 86.55} & {\small 84.83} & {\small 83.79} \\
\hline
\end{tabular}
\caption{\small Repeatability (in $\%$) at $\rho=5$ of Mesh DoG (Gaussian) feature detection algorithm. Average number of detected points: 391.\label{tab:Zaharescu_corr-det2-feat2-pt400}}
\end{table}

\begin{table}
\centering
\begin{tabular}{lccccc}
& \multicolumn{5}{c}{\small\textbf{ Strength}} \\
\cline{2-6}
{\small\textbf{ Transform.}} & {\small\textbf{  1}} & {\small\textbf{ $\leq$2}} & {\small\textbf{ $\leq$3}} & {\small\textbf{ $\leq$4}} & {\small\textbf{ $\leq$5}}\\
\hline
{\small\em Isometry} & {\small 99.75} & {\small 99.83} & {\small 99.85} & {\small 99.87} & {\small 99.87} \\
{\small\em Rasterization} & {\small 98.25} & {\small 98.28} & {\small 98.23} & {\small 98.26} & {\small 98.17} \\
{\small\em Sampling} & {\small 99.62} & {\small 99.61} & {\small 99.52} & {\small 99.40} & {\small 99.52} \\
{\small\em Holes} & {\small 99.75} & {\small 99.76} & {\small 99.70} & {\small 99.67} & {\small 99.61} \\
{\small\em Micro holes} & {\small 99.25} & {\small 99.29} & {\small 99.26} & {\small 99.25} & {\small 99.20} \\
{\small\em Scaling} & {\small 99.92} & {\small 99.92} & {\small 99.92} & {\small 99.92} & {\small 99.92} \\
{\small\em Affine} & {\small 99.73} & {\small 99.83} & {\small 99.83} & {\small 99.83} & {\small 99.83} \\
{\small\em Noise} & {\small 99.94} & {\small 99.92} & {\small 99.92} & {\small 99.92} & {\small 99.91} \\
{\small\em Shot Noise} & {\small 99.74} & {\small 99.75} & {\small 99.74} & {\small 99.71} & {\small 99.71} \\
{\small\em Partial} & {\small 99.91} & {\small 99.88} & {\small 99.90} & {\small 99.82} & {\small 99.85} \\
{\small\em View} & {\small 99.97} & {\small 99.89} & {\small 99.88} & {\small 99.84} & {\small 99.84} \\
\hline
{\small{\textbf{Average}}} & {\small 99.62} & {\small 99.63} & {\small 99.61} & {\small 99.59} & {\small 99.58} \\
\hline
\end{tabular}
\caption{\small Repeatability (in $\%$) at $\rho=5$ of Mesh-Scale DoG (1) feature detection algorithm. Average number of detected points: 3616.\label{tab:keller_sift}}
\end{table}
\begin{table}
\centering
\begin{tabular}{lccccc}
& \multicolumn{5}{c}{\small\textbf{ Strength}} \\
\cline{2-6}
{\small\textbf{ Transform.}} & {\small\textbf{  1}} & {\small\textbf{ $\leq$2}} & {\small\textbf{ $\leq$3}} & {\small\textbf{ $\leq$4}} & {\small\textbf{ $\leq$5}}\\
\hline
{\small\em Isometry} & {\small 97.93} & {\small 98.50} & {\small 98.74} & {\small 98.77} & {\small 98.78} \\
{\small\em Rasterization} & {\small 73.91} & {\small 75.03} & {\small 74.38} & {\small 75.36} & {\small 75.91} \\
{\small\em Sampling} & {\small 93.63} & {\small 91.40} & {\small 89.84} & {\small 88.39} & {\small 89.63} \\
{\small\em Holes} & {\small 94.13} & {\small 92.81} & {\small 91.68} & {\small 90.79} & {\small 90.16} \\
{\small\em Micro holes} & {\small 93.24} & {\small 92.13} & {\small 90.97} & {\small 90.04} & {\small 89.30} \\
{\small\em Scaling} & {\small 98.85} & {\small 98.85} & {\small 98.85} & {\small 98.85} & {\small 98.85} \\
{\small\em Affine} & {\small 96.51} & {\small 96.66} & {\small 96.91} & {\small 96.76} & {\small 96.49} \\
{\small\em Noise} & {\small 94.53} & {\small 95.05} & {\small 95.23} & {\small 95.37} & {\small 95.43} \\
{\small\em Shot Noise} & {\small 94.29} & {\small 93.72} & {\small 93.57} & {\small 93.47} & {\small 93.47} \\
{\small\em Partial} & {\small 98.65} & {\small 98.23} & {\small 98.20} & {\small 97.61} & {\small 97.86} \\
{\small\em View} & {\small 97.77} & {\small 97.53} & {\small 97.55} & {\small 97.28} & {\small 97.05} \\
\hline
{\small{\textbf{Average}}} & {\small 93.95} & {\small 93.63} & {\small 93.27} & {\small 92.97} & {\small 92.99} \\
\hline
\end{tabular}
\caption{\small Repeatability (in $\%$) at $\rho=5$ of Mesh-Scale DoG (2) feature detection algorithm. Average number of detected points: 1538.\label{tab:keller_sift-1}}
\end{table}

\begin{table}
\centering
\begin{tabular}{lccccc}
& \multicolumn{5}{c}{\small\textbf{ Strength}} \\
\cline{2-6}
{\small\textbf{ Transform.}} & {\small\textbf{  1}} & {\small\textbf{ $\leq$2}} & {\small\textbf{ $\leq$3}} & {\small\textbf{ $\leq$4}} & {\small\textbf{ $\leq$5}}\\
\hline
{\small\em Isometry} & {\small 49.18} & {\small 50.20} & {\small 50.38} & {\small 50.81} & {\small 51.03} \\
{\small\em Rasterization} & {\small 31.93} & {\small 31.98} & {\small 32.03} & {\small 31.93} & {\small 31.57} \\
{\small\em Sampling} & {\small 40.59} & {\small 40.09} & {\small 37.97} & {\small 36.71} & {\small 35.82} \\
{\small\em Holes} & {\small 53.71} & {\small 51.78} & {\small 52.31} & {\small 52.33} & {\small 52.29} \\
{\small\em Micro holes} & {\small 50.00} & {\small 50.05} & {\small 50.65} & {\small 51.36} & {\small 51.14} \\
{\small\em Scaling} & {\small 49.73} & {\small 51.44} & {\small 51.80} & {\small 51.77} & {\small 51.21} \\
{\small\em Affine} & {\small 51.38} & {\small 51.27} & {\small 51.33} & {\small 51.36} & {\small 51.23} \\
{\small\em Noise} & {\small 53.91} & {\small 53.94} & {\small 53.17} & {\small 52.57} & {\small 52.51} \\
{\small\em Shot Noise} & {\small 50.77} & {\small 53.56} & {\small 52.88} & {\small 52.32} & {\small 51.96} \\
{\small\em Partial} & {\small 62.26} & {\small 67.72} & {\small 63.65} & {\small 54.13} & {\small 47.32} \\
{\small\em View} & {\small 42.59} & {\small 48.78} & {\small 47.51} & {\small 49.47} & {\small 49.93} \\
\hline
{\small{\textbf{Average}}} & {\small 48.73} & {\small 50.07} & {\small 49.42} & {\small 48.61} & {\small 47.82} \\
\hline
\end{tabular}
\caption{\small Repeatability (in $\%$) at $\rho=5$ of Mesh SIFT (H) feature detection algorithm. Average number of detected points: 2564.\label{tab:Smeets_Dirk_meshSIFT_H}}
\end{table}
\begin{table}
\centering
\begin{tabular}{lccccc}
& \multicolumn{5}{c}{\small\textbf{ Strength}} \\
\cline{2-6}
{\small\textbf{ Transform.}} & {\small\textbf{  1}} & {\small\textbf{ $\leq$2}} & {\small\textbf{ $\leq$3}} & {\small\textbf{ $\leq$4}} & {\small\textbf{ $\leq$5}}\\
\hline
{\small\em Isometry} & {\small 50.11} & {\small 51.63} & {\small 51.74} & {\small 51.62} & {\small 51.86} \\
{\small\em Rasterization} & {\small 34.37} & {\small 35.44} & {\small 34.73} & {\small 34.72} & {\small 34.97} \\
{\small\em Sampling} & {\small 41.13} & {\small 40.03} & {\small 38.64} & {\small 38.63} & {\small 38.18} \\
{\small\em Holes} & {\small 51.79} & {\small 51.67} & {\small 52.53} & {\small 52.21} & {\small 51.63} \\
{\small\em Micro holes} & {\small 51.91} & {\small 52.42} & {\small 52.30} & {\small 52.54} & {\small 52.32} \\
{\small\em Scaling} & {\small 51.86} & {\small 51.47} & {\small 51.45} & {\small 51.68} & {\small 51.58} \\
{\small\em Affine} & {\small 52.25} & {\small 52.34} & {\small 52.33} & {\small 52.08} & {\small 52.09} \\
{\small\em Noise} & {\small 52.70} & {\small 52.74} & {\small 52.72} & {\small 52.33} & {\small 52.34} \\
{\small\em Shot Noise} & {\small 51.62} & {\small 52.17} & {\small 51.91} & {\small 51.65} & {\small 51.76} \\
{\small\em Partial} & {\small 62.46} & {\small 67.70} & {\small 64.49} & {\small 56.13} & {\small 49.17} \\
{\small\em View} & {\small 41.88} & {\small 47.84} & {\small 46.60} & {\small 48.99} & {\small 49.89} \\
\hline
{\small{\textbf{Average}}} & {\small 49.28} & {\small 50.49} & {\small 49.95} & {\small 49.32} & {\small 48.71} \\
\hline
\end{tabular}
\caption{\small Repeatability (in $\%$) at $\rho=5$ of Mesh SIFT (KK) feature detection algorithm. Average number of detected points: 3786.\label{tab:Smeets_Dirk_meshSIFT_KK}}
\end{table}

\begin{table}
\centering
\begin{tabular}{lccccc}
& \multicolumn{5}{c}{\small\textbf{ Strength}} \\
\cline{2-6}
{\small\textbf{ Transform.}} & {\small\textbf{  1}} & {\small\textbf{ $\leq$2}} & {\small\textbf{ $\leq$3}} & {\small\textbf{ $\leq$4}} & {\small\textbf{ $\leq$5}}\\
\hline
{\small\em Isometry} & {\small 99.81} & {\small 99.90} & {\small 99.75} & {\small 99.81} & {\small 99.58} \\
{\small\em Rasterization} & {\small 49.63} & {\small 51.63} & {\small 53.11} & {\small 50.10} & {\small 47.53} \\
{\small\em Sampling} & {\small 97.20} & {\small 98.27} & {\small 98.18} & {\small 97.13} & {\small 87.71} \\
{\small\em Holes} & {\small 100.00} & {\small 99.54} & {\small 99.33} & {\small 99.12} & {\small 98.92} \\
{\small\em Micro holes} & {\small 98.51} & {\small 98.71} & {\small 98.67} & {\small 98.66} & {\small 98.56} \\
{\small\em Scaling} & {\small 100.00} & {\small 100.00} & {\small 100.00} & {\small 99.90} & {\small 99.92} \\
{\small\em Affine} & {\small 99.43} & {\small 99.52} & {\small 98.16} & {\small 95.19} & {\small 93.75} \\
{\small\em Noise} & {\small 98.48} & {\small 97.05} & {\small 95.17} & {\small 94.38} & {\small 93.22} \\
{\small\em Shot Noise} & {\small 95.62} & {\small 93.62} & {\small 92.06} & {\small 91.10} & {\small 90.63} \\
{\small\em Partial} & {\small 100.00} & {\small 99.46} & {\small 99.21} & {\small 99.05} & {\small 99.24} \\
{\small\em View} & {\small 99.75} & {\small 99.77} & {\small 99.85} & {\small 99.66} & {\small 99.73} \\
\hline
{\small{\textbf{Average}}} & {\small 94.40} & {\small 94.32} & {\small 93.95} & {\small 93.10} & {\small 91.71} \\
\hline
\end{tabular}
\caption{\small Repeatability (in $\%$) at $\rho=5$ of Harris3D (ring 1) feature detection algorithm. Average number of detected points: 449.\label{tab:sipiran_ivan_Harris3DRings1}}
\end{table}
\begin{table}
\centering
\begin{tabular}{lccccc}
& \multicolumn{5}{c}{\small\textbf{ Strength}} \\
\cline{2-6}
{\small\textbf{ Transform.}} & {\small\textbf{  1}} & {\small\textbf{ $\leq$2}} & {\small\textbf{ $\leq$3}} & {\small\textbf{ $\leq$4}} & {\small\textbf{ $\leq$5}}\\
\hline
{\small\em Isometry} & {\small 99.81} & {\small 99.90} & {\small 99.87} & {\small 99.90} & {\small 99.77} \\
{\small\em Rasterization} & {\small 49.08} & {\small 51.22} & {\small 52.29} & {\small 47.94} & {\small 44.04} \\
{\small\em Sampling} & {\small 94.00} & {\small 94.67} & {\small 93.11} & {\small 84.83} & {\small 79.87} \\
{\small\em Holes} & {\small 100.00} & {\small 100.00} & {\small 99.94} & {\small 99.72} & {\small 99.63} \\
{\small\em Micro holes} & {\small 99.26} & {\small 99.09} & {\small 99.04} & {\small 98.98} & {\small 98.95} \\
{\small\em Scaling} & {\small 100.00} & {\small 100.00} & {\small 100.00} & {\small 99.86} & {\small 99.89} \\
{\small\em Affine} & {\small 99.43} & {\small 98.95} & {\small 95.62} & {\small 92.33} & {\small 89.98} \\
{\small\em Noise} & {\small 100.00} & {\small 99.24} & {\small 97.97} & {\small 95.86} & {\small 93.45} \\
{\small\em Shot Noise} & {\small 96.95} & {\small 95.71} & {\small 94.67} & {\small 93.90} & {\small 93.45} \\
{\small\em Partial} & {\small 100.00} & {\small 99.46} & {\small 99.35} & {\small 98.55} & {\small 98.84} \\
{\small\em View} & {\small 100.00} & {\small 100.00} & {\small 100.00} & {\small 100.00} & {\small 100.00} \\
\hline
{\small{\textbf{Average}}} & {\small 94.41} & {\small 94.39} & {\small 93.80} & {\small 91.99} & {\small 90.71} \\
\hline
\end{tabular}
\caption{\small Repeatability (in $\%$) at $\rho=5$ of Harris3D (ring 2) feature detection algorithm. Average number of detected points: 449.\label{tab:sipiran_ivan_Harris3DRings2}}
\end{table}
\begin{table}
\centering
\begin{tabular}{lccccc}
& \multicolumn{5}{c}{\small\textbf{ Strength}} \\
\cline{2-6}
{\small\textbf{ Transform.}} & {\small\textbf{  1}} & {\small\textbf{ $\leq$2}} & {\small\textbf{ $\leq$3}} & {\small\textbf{ $\leq$4}} & {\small\textbf{ $\leq$5}}\\
\hline
{\small\em Isometry} & {\small 96.57} & {\small 98.10} & {\small 98.60} & {\small 98.81} & {\small 98.86} \\
{\small\em Rasterization} & {\small 52.39} & {\small 55.55} & {\small 60.09} & {\small 56.80} & {\small 53.32} \\
{\small\em Sampling} & {\small 90.80} & {\small 95.07} & {\small 94.71} & {\small 92.03} & {\small 91.63} \\
{\small\em Holes} & {\small 99.44} & {\small 99.08} & {\small 98.71} & {\small 97.96} & {\small 97.50} \\
{\small\em Micro holes} & {\small 98.70} & {\small 98.81} & {\small 98.56} & {\small 98.44} & {\small 98.28} \\
{\small\em Scaling} & {\small 99.43} & {\small 99.33} & {\small 99.11} & {\small 99.05} & {\small 98.78} \\
{\small\em Affine} & {\small 96.38} & {\small 92.95} & {\small 87.17} & {\small 81.71} & {\small 78.78} \\
{\small\em Noise} & {\small 50.48} & {\small 51.90} & {\small 52.57} & {\small 54.10} & {\small 55.24} \\
{\small\em Shot Noise} & {\small 93.14} & {\small 90.57} & {\small 89.40} & {\small 88.38} & {\small 87.62} \\
{\small\em Partial} & {\small 99.33} & {\small 98.46} & {\small 98.61} & {\small 95.95} & {\small 96.76} \\
{\small\em View} & {\small 99.26} & {\small 98.60} & {\small 98.73} & {\small 98.37} & {\small 98.43} \\
\hline
{\small{\textbf{Average}}} & {\small 88.72} & {\small 88.95} & {\small 88.75} & {\small 87.42} & {\small 86.84} \\
\hline
\end{tabular}
\caption{\small Repeatability (in $\%$) at $\rho=5$ of Harris3D (Adaptive) feature detection algorithm. Average number of detected points: 449.\label{tab:sipiran_ivan_Harris3DAdaptive}}
\end{table}

\subsection{Region feature detectors.}

\begin{figure*}[t]
    \includegraphics[width=\linewidth]{Legend_H-eps-converted-to.pdf} \\
     \begin{tabular}{cc}
        \includegraphics[width=\columnwidth]{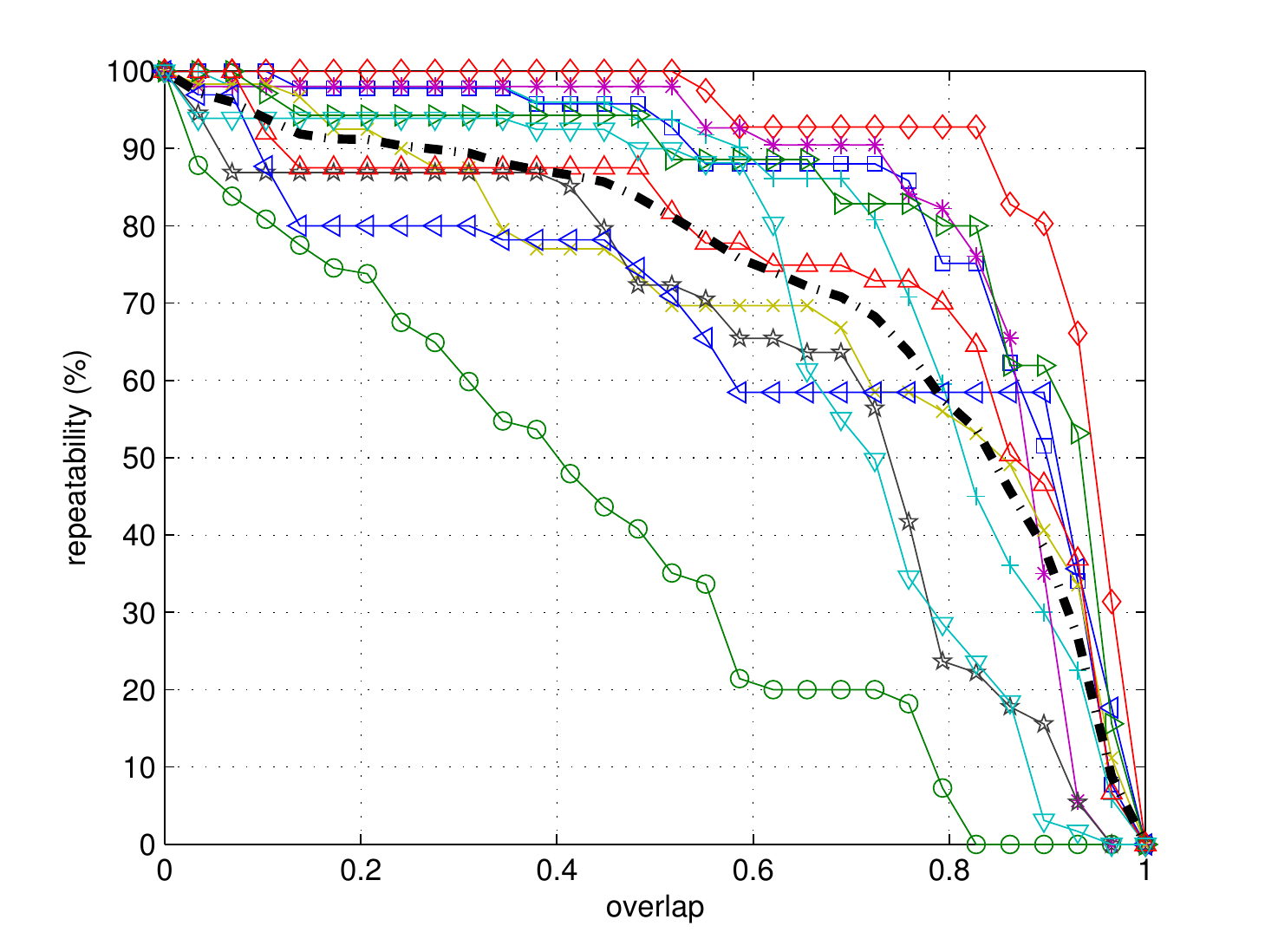} &
        \includegraphics[width=\columnwidth]{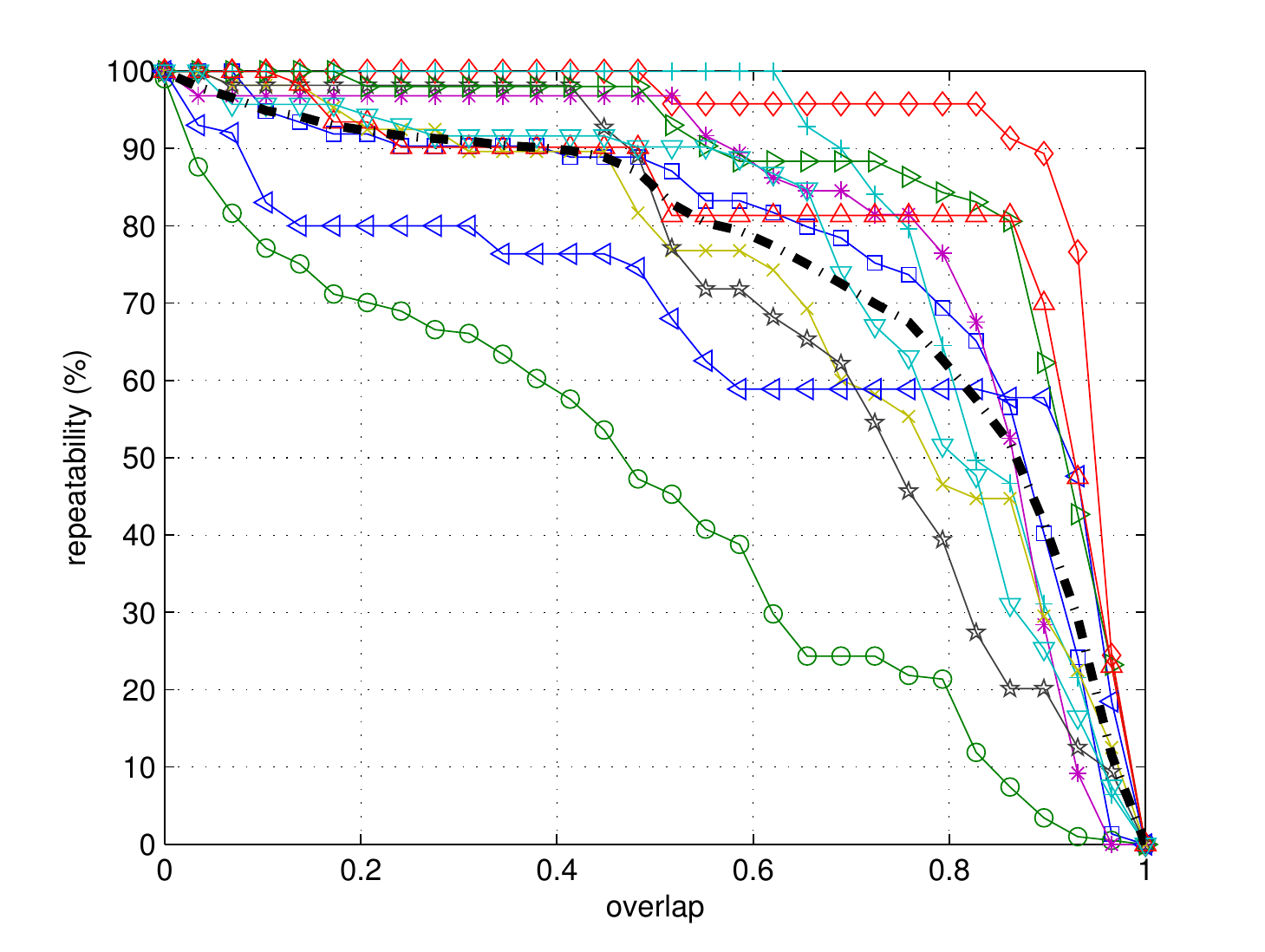} \\
        {\small Shape MSER (EW 1/CT)} & {\small Shape MSER (EW 1/HKS)}\\                
	\multicolumn{2}{c}{\includegraphics[width=\columnwidth]{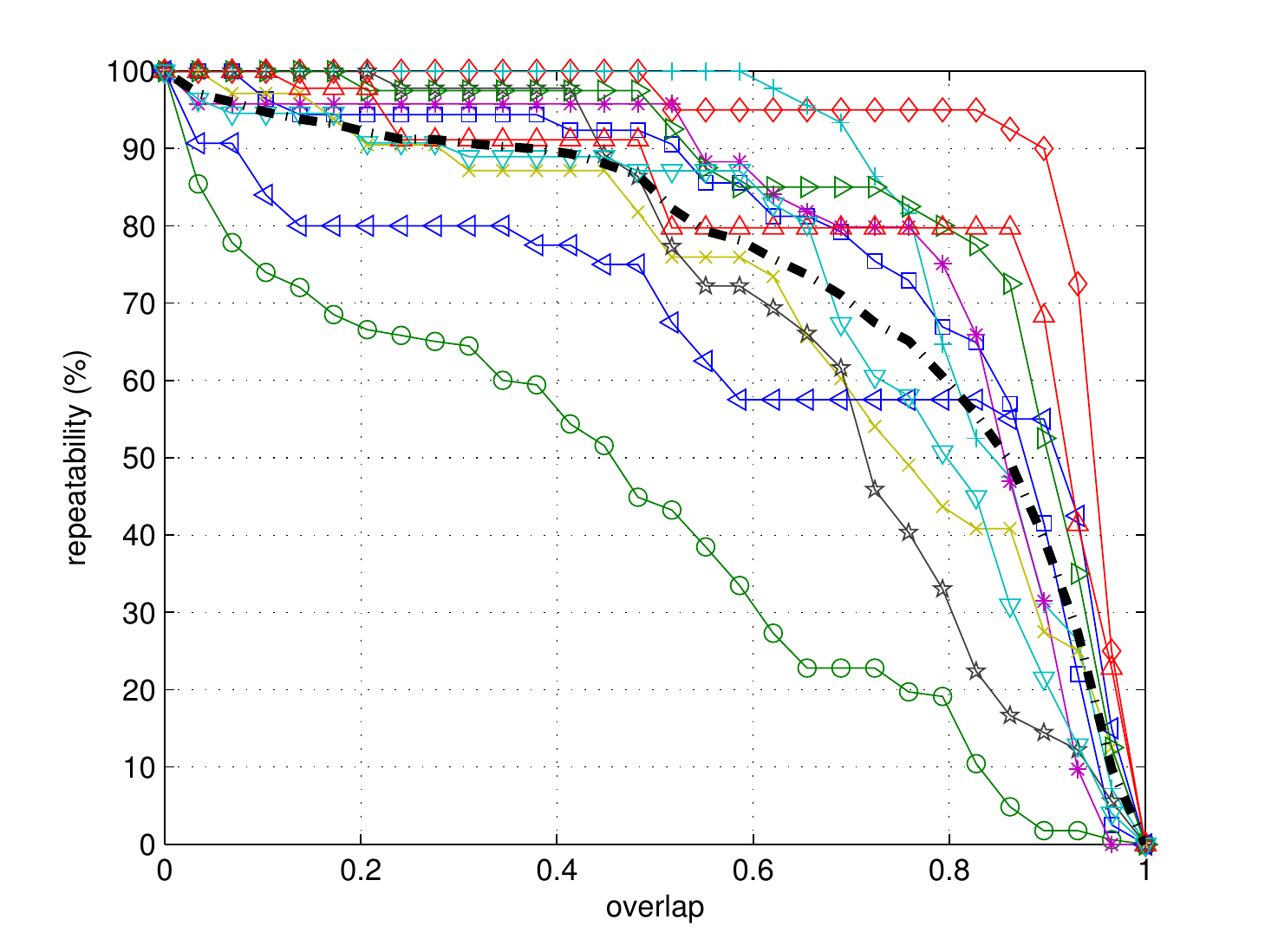}} \\
 	\multicolumn{2}{c}{\small Shape MSER (VW HKS)} \\          
    \end{tabular}
\caption{Repeatability ($\%$) vs overlap of region feature detectors broken down according to different transformation classes. }
\label{fig_repeatability_region}
\end{figure*}

Tables~\ref{tab:shapemser_rep_ew1_hks}--\ref{tab:shapemser_rep_vw} show the repeatability of different point descriptors at fixed overlap of $0.7$, broken down according to transformation classes and strengths. 
Figure~\ref{fig_repeatability_region} shows the repeatability of region feature detectors as function of overlap varying from $0$ to $1$. 

\begin{table}
\centering
\begin{tabular}{lccccc}
& \multicolumn{5}{c}{\small\textbf{ Strength}} \\
\cline{2-6}
{\small\textbf{ Transform.}} & {\small\textbf{  1}} & {\small\textbf{ $\leq$2}} & {\small\textbf{ $\leq$3}} & {\small\textbf{ $\leq$4}} & {\small\textbf{ $\leq$5}}\\
\hline
{\small\em Isometry} & {\small 100.00} & {\small 100.00} & {\small 96.30} & {\small 94.72} & {\small 95.78}\\
{\small\em Rasterization} & {\small 46.67} & {\small 64.24} & {\small 66.64} & {\small 61.34} & {\small 54.53}\\
{\small\em Sampling} & {\small 100.00} & {\small 100.00} & {\small 98.15} & {\small 73.61} & {\small 58.89}\\
{\small\em Holes} & {\small 90.00} & {\small 45.00} & {\small 30.00} & {\small 22.50} & {\small 24.34}\\
{\small\em Micro holes} & {\small 100.00} & {\small 100.00} & {\small 100.00} & {\small 96.15} & {\small 84.07}\\
{\small\em Scaling} & {\small 87.50} & {\small 87.50} & {\small 89.29} & {\small 86.96} & {\small 88.32}\\
{\small\em Affine} & {\small 93.33} & {\small 90.42} & {\small 78.46} & {\small 76.70} & {\small 75.21}\\
{\small\em Noise} & {\small 83.33} & {\small 87.12} & {\small 87.71} & {\small 84.53} & {\small 81.47}\\
{\small\em Shot Noise} & {\small 92.86} & {\small 90.18} & {\small 83.45} & {\small 82.90} & {\small 81.32}\\
{\small\em Partial} & {\small 75.00} & {\small 77.50} & {\small 57.73} & {\small 62.05} & {\small 58.21}\\
{\small\em View} & {\small 20.00} & {\small 60.00} & {\small 61.43} & {\small 69.51} & {\small 67.04}\\
\hline
{\small{\textbf{Average}}} & {\small 80.79} & {\small 82.00} & {\small 77.20} & {\small 73.73} & {\small 69.92}\\
\hline
\end{tabular}
\caption{\small Repeatability (in $\%$) at {\em overlap $\geq$ 0.7} of Shape MSER (EW 1/HKS) region detector algorithm. Average number of detected regions: 12.36.\label{tab:shapemser_rep_ew1_hks}}
\end{table}
\begin{table}
\centering
\begin{tabular}{lccccc}
& \multicolumn{5}{c}{\small\textbf{ Strength}} \\
\cline{2-6}
{\small\textbf{ Transform.}} & {\small\textbf{  1}} & {\small\textbf{ $\leq$2}} & {\small\textbf{ $\leq$3}} & {\small\textbf{ $\leq$4}} & {\small\textbf{ $\leq$5}}\\
\hline
{\small\em Isometry} & {\small 88.89} & {\small 94.44} & {\small 92.13} & {\small 94.10} & {\small 92.78}\\
{\small\em Rasterization} & {\small 77.78} & {\small 78.17} & {\small 78.78} & {\small 65.91} & {\small 56.36}\\
{\small\em Sampling} & {\small 92.31} & {\small 96.15} & {\small 97.44} & {\small 73.08} & {\small 58.46}\\
{\small\em Holes} & {\small 100.00} & {\small 50.00} & {\small 33.33} & {\small 25.00} & {\small 20.00}\\
{\small\em Micro holes} & {\small 100.00} & {\small 100.00} & {\small 96.30} & {\small 90.97} & {\small 80.78}\\
{\small\em Scaling} & {\small 14.29} & {\small 57.14} & {\small 71.43} & {\small 78.57} & {\small 82.86}\\
{\small\em Affine} & {\small 90.91} & {\small 88.31} & {\small 87.08} & {\small 87.81} & {\small 88.03}\\
{\small\em Noise} & {\small 88.89} & {\small 86.11} & {\small 90.74} & {\small 93.06} & {\small 90.44}\\
{\small\em Shot Noise} & {\small 100.00} & {\small 95.83} & {\small 78.17} & {\small 81.13} & {\small 72.90}\\
{\small\em Partial} & {\small 83.33} & {\small 84.52} & {\small 64.68} & {\small 63.10} & {\small 58.48}\\
{\small\em View} & {\small 11.11} & {\small 38.89} & {\small 37.83} & {\small 46.55} & {\small 49.74}\\
\hline
{\small{\textbf{Average}}} & {\small 77.05} & {\small 79.05} & {\small 75.26} & {\small 72.66} & {\small 68.26}\\
\hline
\end{tabular}
\caption{\small Repeatability (in $\%$) at {\em overlap $\geq$ 0.7} of Shape MSER (EW 1/CT) region detector algorithm. Average number of detected regions: 8.85.\label{tab:shapemser_rep_ew1_ct}}
\end{table}
\begin{table}
\centering
\begin{tabular}{lccccc}
& \multicolumn{5}{c}{\small\textbf{ Strength}} \\
\cline{2-6}
{\small\textbf{ Transform.}} & {\small\textbf{  1}} & {\small\textbf{ $\leq$2}} & {\small\textbf{ $\leq$3}} & {\small\textbf{ $\leq$4}} & {\small\textbf{ $\leq$5}}\\
\hline
{\small\em Isometry} & {\small 100.00} & {\small 100.00} & {\small 95.83} & {\small 93.75} & {\small 95.00}\\
{\small\em Rasterization} & {\small 55.56} & {\small 61.11} & {\small 57.41} & {\small 50.20} & {\small 45.87}\\
{\small\em Sampling} & {\small 100.00} & {\small 100.00} & {\small 95.83} & {\small 71.88} & {\small 57.50}\\
{\small\em Holes} & {\small 87.50} & {\small 43.75} & {\small 29.17} & {\small 21.88} & {\small 22.79}\\
{\small\em Micro holes} & {\small 100.00} & {\small 100.00} & {\small 100.00} & {\small 96.88} & {\small 86.39}\\
{\small\em Scaling} & {\small 87.50} & {\small 87.50} & {\small 87.50} & {\small 84.38} & {\small 85.00}\\
{\small\em Affine} & {\small 87.50} & {\small 87.50} & {\small 76.52} & {\small 72.39} & {\small 75.41}\\
{\small\em Noise} & {\small 87.50} & {\small 87.50} & {\small 87.50} & {\small 82.29} & {\small 79.83}\\
{\small\em Shot Noise} & {\small 87.50} & {\small 82.64} & {\small 81.02} & {\small 80.21} & {\small 79.72}\\
{\small\em Partial} & {\small 75.00} & {\small 75.00} & {\small 59.52} & {\small 63.39} & {\small 54.05}\\
{\small\em View} & {\small 12.50} & {\small 56.25} & {\small 55.68} & {\small 64.26} & {\small 60.50}\\
\hline
{\small{\textbf{Average}}} & {\small 80.05} & {\small 80.11} & {\small 75.09} & {\small 71.04} & {\small 67.46}\\
\hline
\end{tabular}
\caption{\small Repeatability (in $\%$) at {\em overlap $\geq$ 0.7} of Shape MSER (VW HKS) region detector algorithm. Average number of detected regions: 9.25.\label{tab:shapemser_rep_vw}}
\end{table}

\subsection{Point feature descriptors}

\begin{figure*}[t]
    \includegraphics[width=\linewidth]{Legend_H-eps-converted-to.pdf} \\
     \begin{tabular}{cc}
        \includegraphics[width=\columnwidth]{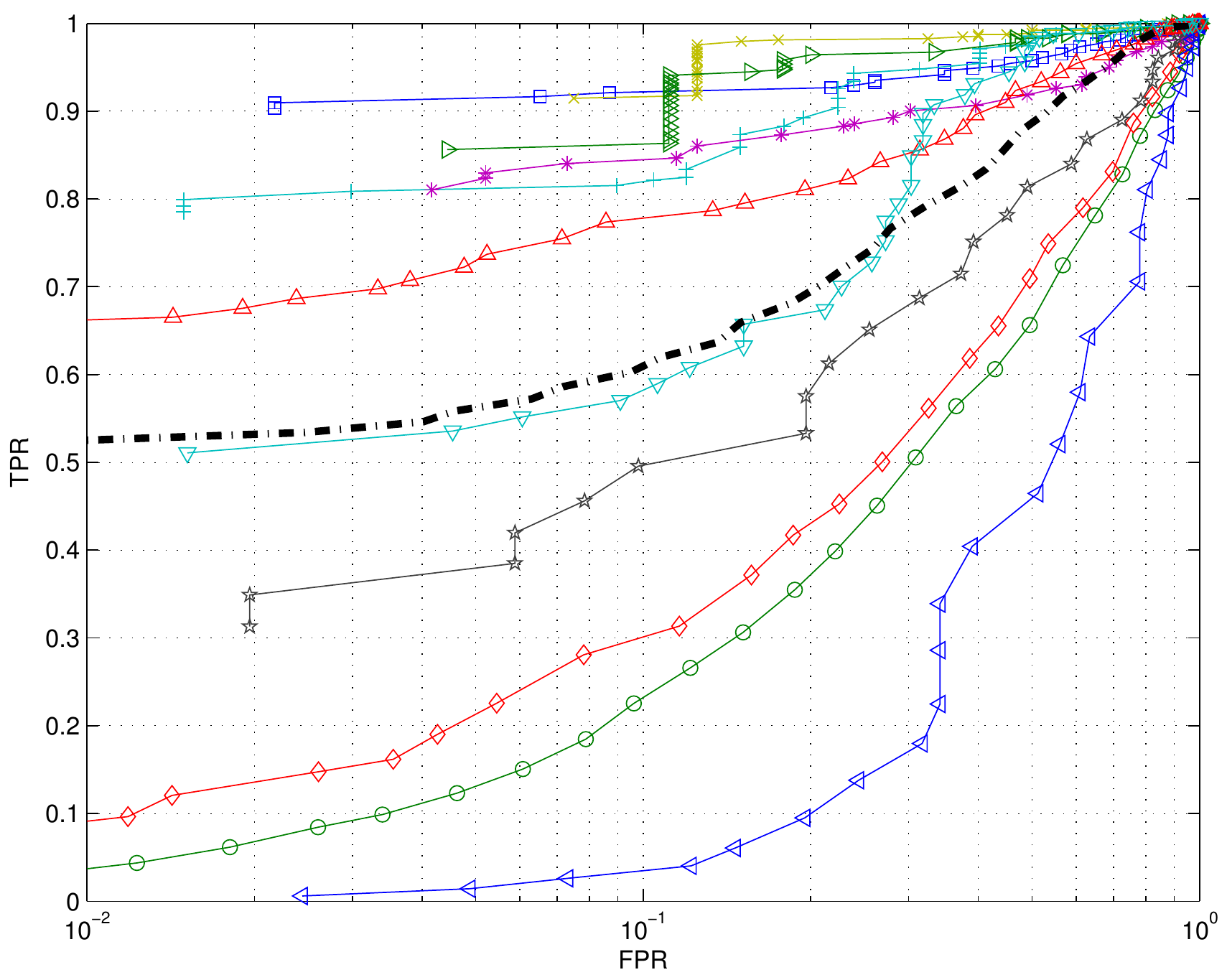} &
        \includegraphics[width=\columnwidth]{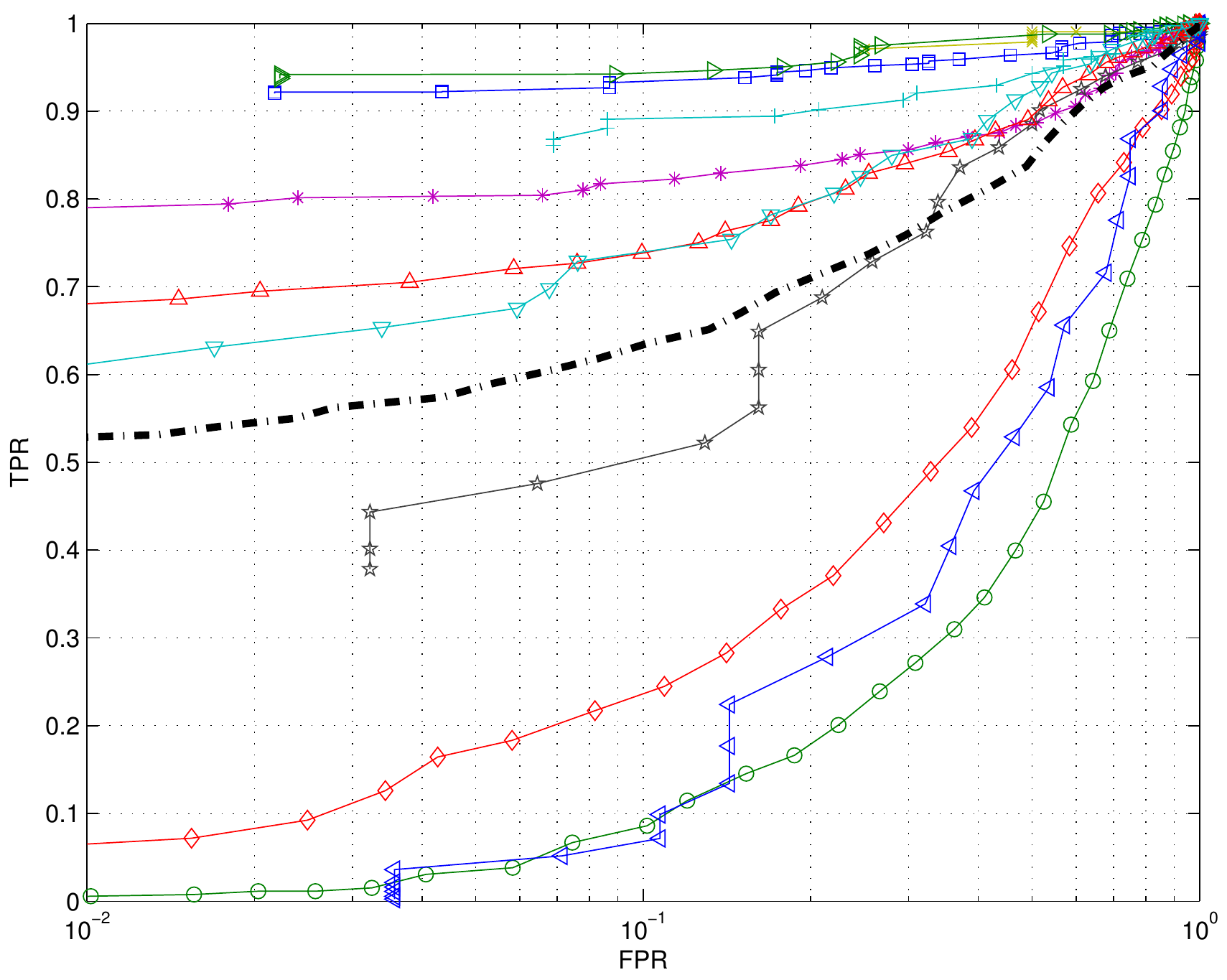} \\
        {\small Mesh HoG (mean)} & {\small Mesh HoG (2)}\\                
        \includegraphics[width=\columnwidth]{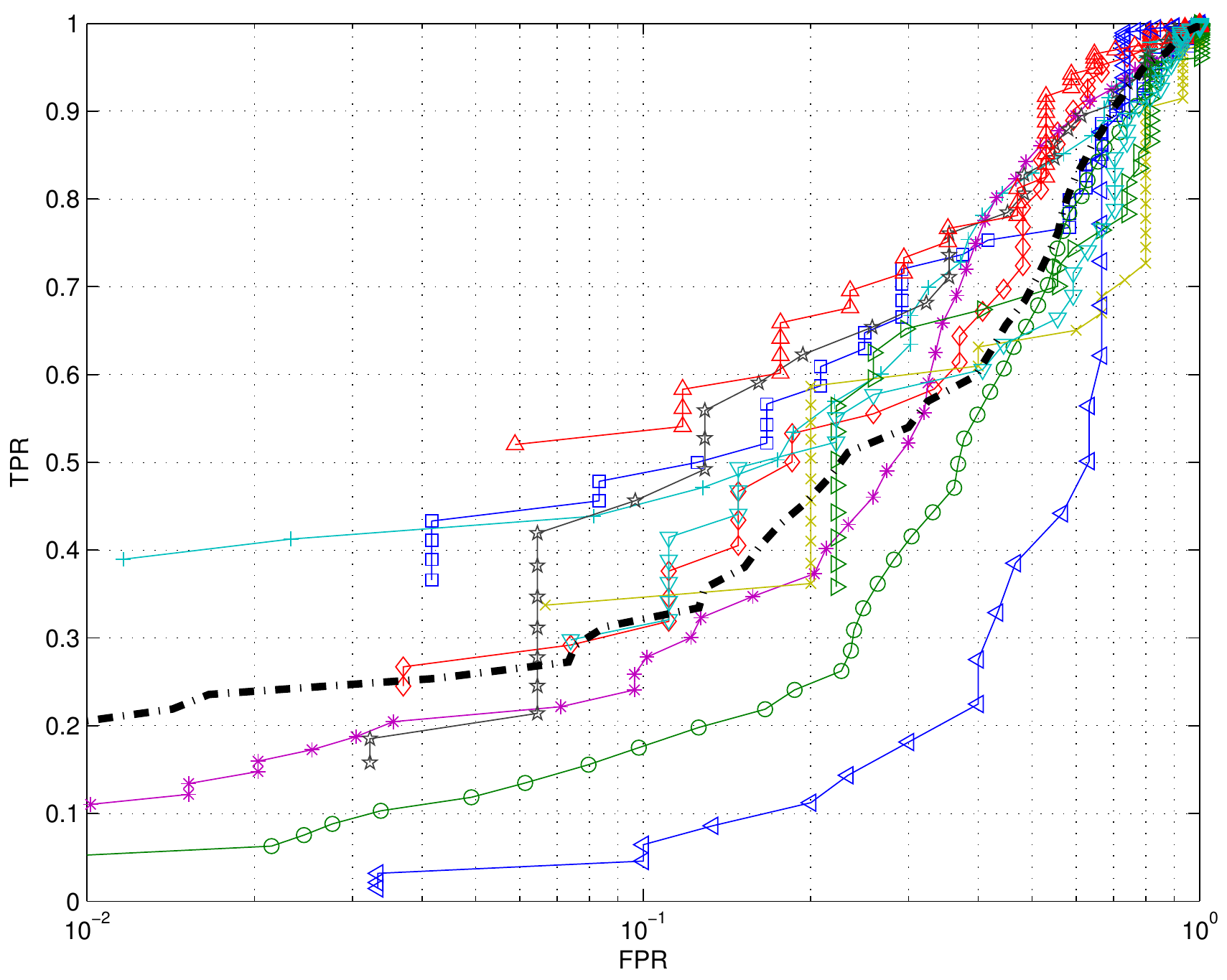} &
        \includegraphics[width=\columnwidth]{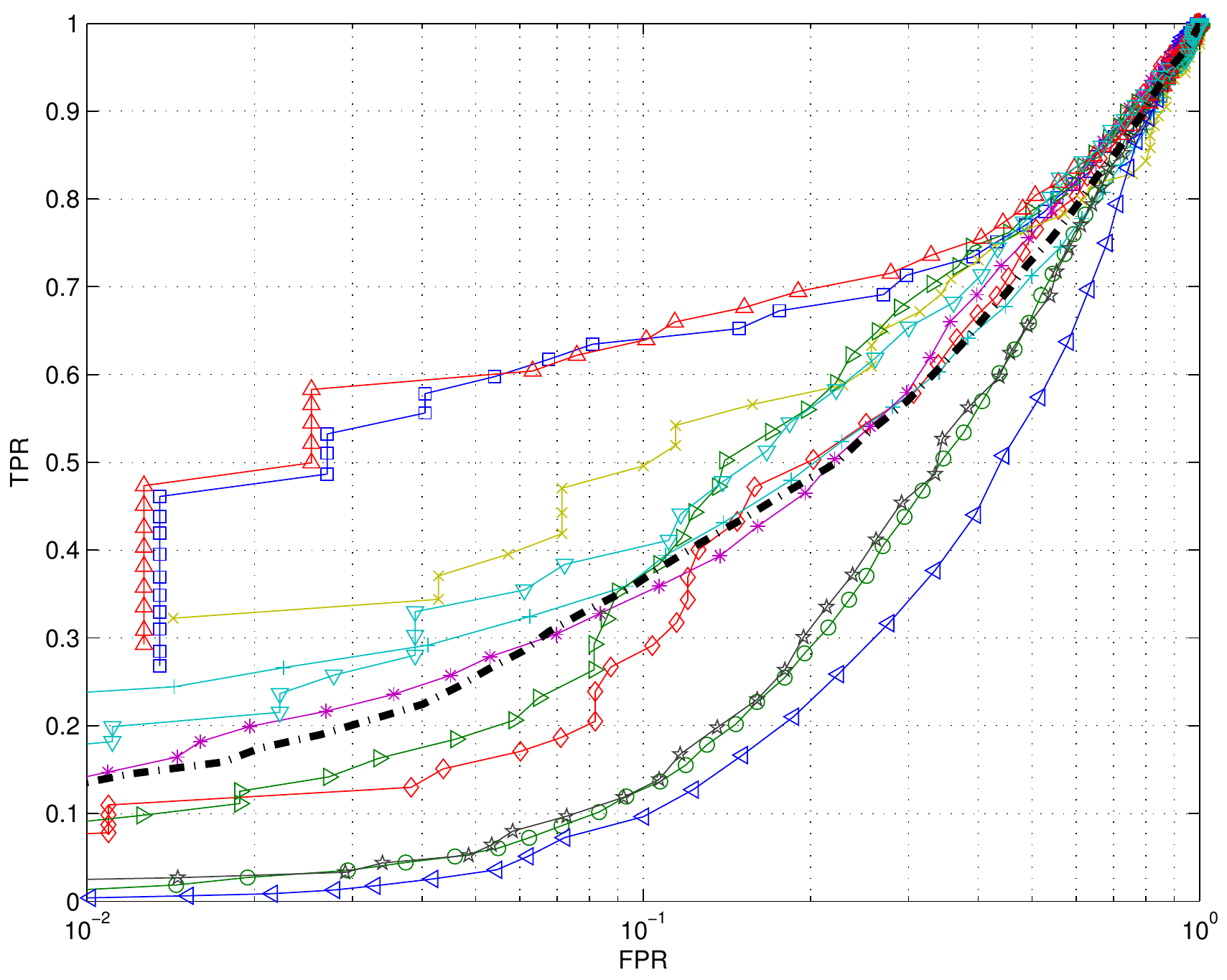} \\
        {\small Local depth SIFT (1)} & {\small Local depth SIFT (2)}\\                
        \includegraphics[width=\columnwidth]{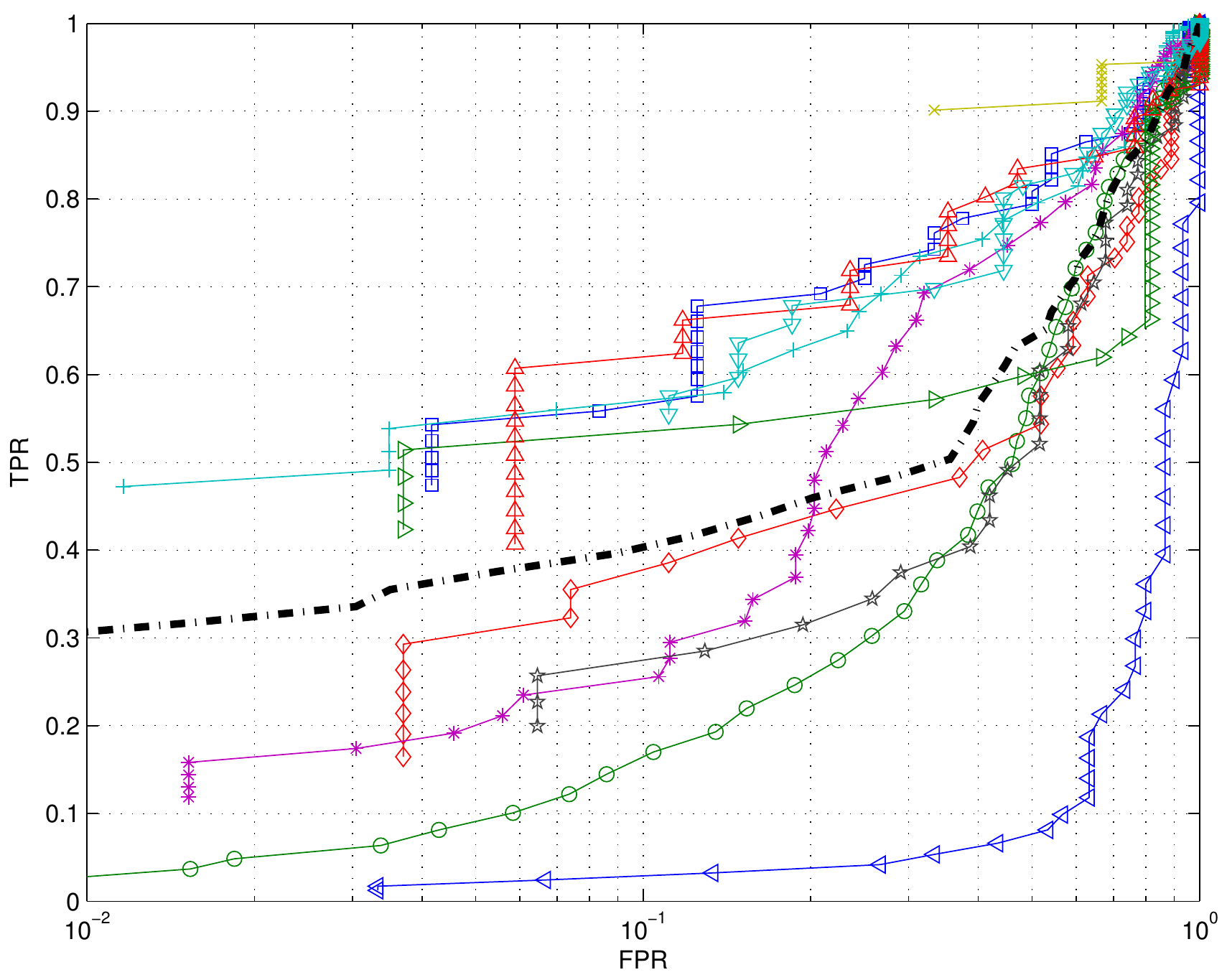} &
        \includegraphics[width=\columnwidth]{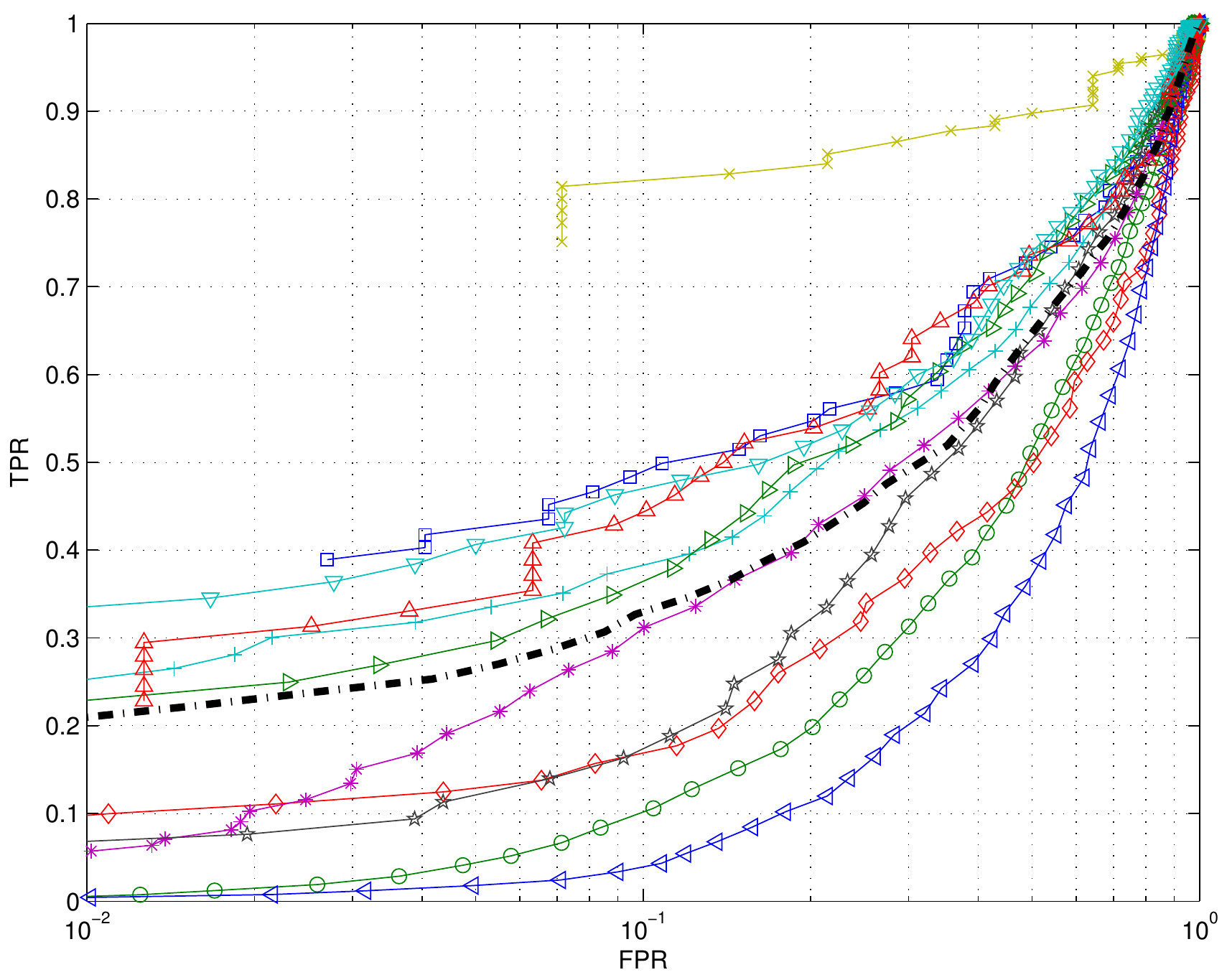} \\
        {\small Scale invariant Spin Image (1)} & {\small Scale invariant Spin Image (2)}\\                
    \end{tabular}
\caption{ROC curves of point feature descriptors broken down according to different transformation classes. }
\label{fig_roc_point}
\end{figure*}

Tables~\ref{tab:Zaharescu_corr-det1-feat1-pt400.desc}--\ref{tab:keller_spin-1.desc} show the performance of different point feature description algorithms, in terms of  average normalized $L_2$ distance between corresponding descriptors. Smaller numbers correspond to better performance.  
Figure~\ref{fig_roc_point} shows the ROC curves of different point feature descriptors, using a fixed value of $\rho=5$. Higher values of the vertical axis at a fixed point on the horizontal axis are indication of better performance.

\begin{table}
\centering
\begin{tabular}{lccccc}
& \multicolumn{5}{c}{\small\textbf{ Strength}} \\
\cline{2-6}
{\small\textbf{ Transform.}} & {\small\textbf{  1}} & {\small\textbf{ $\leq$2}} & {\small\textbf{ $\leq$3}} & {\small\textbf{ $\leq$4}} & {\small\textbf{ $\leq$5}}\\
\hline
{\small\em Isometry} & {\small 0.17} & {\small 0.17} & {\small 0.19} & {\small 0.19} & {\small 0.21} \\
{\small\em Rasterization} & {\small 1.00} & {\small 1.00} & {\small 0.99} & {\small 0.99} & {\small 0.99} \\
{\small\em Sampling} & {\small 0.93} & {\small 0.95} & {\small 0.96} & {\small 0.97} & {\small 0.98} \\
{\small\em Holes} & {\small 0.23} & {\small 0.24} & {\small 0.26} & {\small 0.29} & {\small 0.32} \\
{\small\em Micro holes} & {\small 0.30} & {\small 0.31} & {\small 0.33} & {\small 0.34} & {\small 0.35} \\
{\small\em Scaling} & {\small 0.20} & {\small 0.20} & {\small 0.20} & {\small 0.20} & {\small 0.20} \\
{\small\em Affine} & {\small 0.52} & {\small 0.63} & {\small 0.68} & {\small 0.72} & {\small 0.75} \\
{\small\em Noise} & {\small 1.00} & {\small 0.99} & {\small 0.99} & {\small 0.99} & {\small 0.99} \\
{\small\em Shot Noise} & {\small 0.23} & {\small 0.24} & {\small 0.25} & {\small 0.26} & {\small 0.27} \\
{\small\em Partial} & {\small 0.30} & {\small 0.36} & {\small 0.37} & {\small 0.46} & {\small 0.49} \\
{\small\em View} & {\small 0.64} & {\small 0.59} & {\small 0.61} & {\small 0.60} & {\small 0.60} \\
\hline
{\small{\textbf{Average}}} & {\small 0.50} & {\small 0.52} & {\small 0.53} & {\small 0.55} & {\small 0.56} \\
\hline
\end{tabular}
\caption{\small Quality of Mesh HoG feature description algorithm (average normalized $L_2$ distance between descriptors at corresponding points) on feature points detected using Mesh DoG (mean). Average number of points: 392.\label{tab:Zaharescu_corr-det1-feat1-pt400.desc}}
\end{table}
\begin{table}
\centering
\begin{tabular}{lccccc}
& \multicolumn{5}{c}{\small\textbf{ Strength}} \\
\cline{2-6}
{\small\textbf{ Transform.}} & {\small\textbf{  1}} & {\small\textbf{ $\leq$2}} & {\small\textbf{ $\leq$3}} & {\small\textbf{ $\leq$4}} & {\small\textbf{ $\leq$5}}\\
\hline
{\small\em Isometry} & {\small 0.15} & {\small 0.15} & {\small 0.17} & {\small 0.17} & {\small 0.19} \\
{\small\em Rasterization} & {\small 0.98} & {\small 0.98} & {\small 0.99} & {\small 0.99} & {\small 0.99} \\
{\small\em Sampling} & {\small 0.95} & {\small 0.97} & {\small 0.97} & {\small 0.98} & {\small 0.98} \\
{\small\em Holes} & {\small 0.21} & {\small 0.23} & {\small 0.27} & {\small 0.30} & {\small 0.34} \\
{\small\em Micro holes} & {\small 0.29} & {\small 0.30} & {\small 0.32} & {\small 0.34} & {\small 0.35} \\
{\small\em Scaling} & {\small 0.19} & {\small 0.19} & {\small 0.19} & {\small 0.19} & {\small 0.19} \\
{\small\em Affine} & {\small 0.58} & {\small 0.66} & {\small 0.71} & {\small 0.74} & {\small 0.78} \\
{\small\em Noise} & {\small 0.98} & {\small 0.99} & {\small 0.99} & {\small 0.99} & {\small 0.99} \\
{\small\em Shot Noise} & {\small 0.22} & {\small 0.23} & {\small 0.24} & {\small 0.25} & {\small 0.26} \\
{\small\em Partial} & {\small 0.32} & {\small 0.40} & {\small 0.40} & {\small 0.48} & {\small 0.51} \\
{\small\em View} & {\small 0.69} & {\small 0.62} & {\small 0.65} & {\small 0.65} & {\small 0.63} \\
\hline
{\small{\textbf{Average}}} & {\small 0.51} & {\small 0.52} & {\small 0.54} & {\small 0.55} & {\small 0.56} \\
\hline
\end{tabular}
\caption{\small Quality of Mesh HoG feature description algorithm (average normalized $L_2$ distance between descriptors at corresponding points) on feature points detected using Mesh DoG (Gaussian). Average number of points: 391.\label{tab:Zaharescu_corr-det2-feat2-pt400}}
\end{table}

\begin{table}
\centering
\begin{tabular}{lccccc}
& \multicolumn{5}{c}{\small\textbf{ Strength}} \\
\cline{2-6}
{\small\textbf{ Transform.}} & {\small\textbf{  1}} & {\small\textbf{ $\leq$2}} & {\small\textbf{ $\leq$3}} & {\small\textbf{ $\leq$4}} & {\small\textbf{ $\leq$5}}\\
\hline
{\small\em Isometry} & {\small 0.63} & {\small 0.56} & {\small 0.58} & {\small 0.59} & {\small 0.59} \\
{\small\em Rasterization} & {\small 0.94} & {\small 0.94} & {\small 0.94} & {\small 0.94} & {\small 0.94} \\
{\small\em Sampling} & {\small 0.74} & {\small 0.77} & {\small 0.81} & {\small 0.84} & {\small 0.87} \\
{\small\em Holes} & {\small 0.71} & {\small 0.73} & {\small 0.75} & {\small 0.77} & {\small 0.78} \\
{\small\em Micro holes} & {\small 0.70} & {\small 0.74} & {\small 0.76} & {\small 0.78} & {\small 0.80} \\
{\small\em Scaling} & {\small 0.68} & {\small 0.70} & {\small 0.72} & {\small 0.74} & {\small 0.75} \\
{\small\em Affine} & {\small 0.87} & {\small 0.88} & {\small 0.90} & {\small 0.91} & {\small 0.93} \\
{\small\em Noise} & {\small 0.95} & {\small 0.96} & {\small 0.97} & {\small 0.98} & {\small 0.98} \\
{\small\em Shot Noise} & {\small 0.70} & {\small 0.73} & {\small 0.76} & {\small 0.78} & {\small 0.80} \\
{\small\em Partial} & {\small 0.46} & {\small 0.47} & {\small 0.53} & {\small 0.57} & {\small 0.59} \\
{\small\em View} & {\small 0.76} & {\small 0.74} & {\small 0.75} & {\small 0.74} & {\small 0.74} \\
\hline
{\small{\textbf{Average}}} & {\small 0.74} & {\small 0.75} & {\small 0.77} & {\small 0.79} & {\small 0.80} \\
\hline
\end{tabular}
\caption{\small Quality of Local depth SIFT feature description algorithm (average normalized $L_2$ distance between descriptors at corresponding points) on feature points detected using Mesh-Scale DoG (1). Average number of points: 3616.\label{tab:keller_sift}}
\end{table}
\begin{table}
\centering
\begin{tabular}{lccccc}
& \multicolumn{5}{c}{\small\textbf{ Strength}} \\
\cline{2-6}
{\small\textbf{ Transform.}} & {\small\textbf{  1}} & {\small\textbf{ $\leq$2}} & {\small\textbf{ $\leq$3}} & {\small\textbf{ $\leq$4}} & {\small\textbf{ $\leq$5}}\\
\hline
{\small\em Isometry} & {\small 0.67} & {\small 0.61} & {\small 0.64} & {\small 0.65} & {\small 0.64} \\
{\small\em Rasterization} & {\small 0.95} & {\small 0.96} & {\small 0.96} & {\small 0.96} & {\small 0.96} \\
{\small\em Sampling} & {\small 0.79} & {\small 0.81} & {\small 0.85} & {\small 0.87} & {\small 0.90} \\
{\small\em Holes} & {\small 0.79} & {\small 0.82} & {\small 0.84} & {\small 0.86} & {\small 0.87} \\
{\small\em Micro holes} & {\small 0.78} & {\small 0.80} & {\small 0.83} & {\small 0.84} & {\small 0.86} \\
{\small\em Scaling} & {\small 0.71} & {\small 0.72} & {\small 0.74} & {\small 0.75} & {\small 0.76} \\
{\small\em Affine} & {\small 0.89} & {\small 0.89} & {\small 0.91} & {\small 0.92} & {\small 0.94} \\
{\small\em Noise} & {\small 0.95} & {\small 0.96} & {\small 0.97} & {\small 0.98} & {\small 0.98} \\
{\small\em Shot Noise} & {\small 0.79} & {\small 0.81} & {\small 0.82} & {\small 0.83} & {\small 0.84} \\
{\small\em Partial} & {\small 0.50} & {\small 0.51} & {\small 0.57} & {\small 0.61} & {\small 0.63} \\
{\small\em View} & {\small 0.84} & {\small 0.82} & {\small 0.83} & {\small 0.83} & {\small 0.83} \\
\hline
{\small{\textbf{Average}}} & {\small 0.79} & {\small 0.79} & {\small 0.81} & {\small 0.83} & {\small 0.84} \\
\hline
\end{tabular}
\caption{\small Quality of Local depth SIFT feature description algorithm (average normalized $L_2$ distance between descriptors at corresponding points) on feature points detected using Mesh-Scale DoG (2). Average number of points: 1538.\label{tab:keller_sift-1}}
\end{table}

\begin{table}
\centering
\begin{tabular}{lccccc}
& \multicolumn{5}{c}{\small\textbf{ Strength}} \\
\cline{2-6}
{\small\textbf{ Transform.}} & {\small\textbf{  1}} & {\small\textbf{ $\leq$2}} & {\small\textbf{ $\leq$3}} & {\small\textbf{ $\leq$4}} & {\small\textbf{ $\leq$5}}\\
\hline
{\small\em Isometry} & {\small 0.26} & {\small 0.52} & {\small 0.61} & {\small 0.52} & {\small 0.57} \\
{\small\em Rasterization} & {\small 0.88} & {\small 0.87} & {\small 0.87} & {\small 0.87} & {\small 0.87} \\
{\small\em Sampling} & {\small 0.62} & {\small 0.66} & {\small 0.70} & {\small 0.75} & {\small 0.79} \\
{\small\em Holes} & {\small 0.36} & {\small 0.39} & {\small 0.42} & {\small 0.45} & {\small 0.47} \\
{\small\em Micro holes} & {\small 0.84} & {\small 0.85} & {\small 0.86} & {\small 0.87} & {\small 0.87} \\
{\small\em Scaling} & {\small 0.24} & {\small 0.24} & {\small 0.24} & {\small 0.24} & {\small 0.24} \\
{\small\em Affine} & {\small 0.60} & {\small 0.74} & {\small 0.75} & {\small 0.76} & {\small 0.79} \\
{\small\em Noise} & {\small 0.96} & {\small 0.97} & {\small 0.98} & {\small 0.98} & {\small 0.99} \\
{\small\em Shot Noise} & {\small 0.41} & {\small 0.48} & {\small 0.52} & {\small 0.56} & {\small 0.60} \\
{\small\em Partial} & {\small 0.78} & {\small 0.77} & {\small 0.78} & {\small 0.69} & {\small 0.62} \\
{\small\em View} & {\small 0.50} & {\small 0.45} & {\small 0.46} & {\small 0.46} & {\small 0.45} \\
\hline
{\small{\textbf{Average}}} & {\small 0.58} & {\small 0.63} & {\small 0.65} & {\small 0.65} & {\small 0.66} \\
\hline
\end{tabular}
\caption{\small Quality of Scale invariant Spin Image feature description algorithm (average normalized $L_2$ distance between descriptors at corresponding points) on feature points detected using Mesh-Scale DoG (1). Average number of points: 3616.\label{tab:keller_spin}}
\end{table}
\begin{table}
\centering
\begin{tabular}{lccccc}
& \multicolumn{5}{c}{\small\textbf{ Strength}} \\
\cline{2-6}
{\small\textbf{ Transform.}} & {\small\textbf{  1}} & {\small\textbf{ $\leq$2}} & {\small\textbf{ $\leq$3}} & {\small\textbf{ $\leq$4}} & {\small\textbf{ $\leq$5}}\\
\hline
{\small\em Isometry} & {\small 0.29} & {\small 0.55} & {\small 0.64} & {\small 0.55} & {\small 0.60} \\
{\small\em Rasterization} & {\small 0.91} & {\small 0.90} & {\small 0.89} & {\small 0.89} & {\small 0.90} \\
{\small\em Sampling} & {\small 0.67} & {\small 0.71} & {\small 0.74} & {\small 0.78} & {\small 0.82} \\
{\small\em Holes} & {\small 0.50} & {\small 0.55} & {\small 0.59} & {\small 0.62} & {\small 0.65} \\
{\small\em Micro holes} & {\small 0.89} & {\small 0.90} & {\small 0.90} & {\small 0.91} & {\small 0.92} \\
{\small\em Scaling} & {\small 0.26} & {\small 0.26} & {\small 0.26} & {\small 0.26} & {\small 0.26} \\
{\small\em Affine} & {\small 0.64} & {\small 0.77} & {\small 0.77} & {\small 0.78} & {\small 0.81} \\
{\small\em Noise} & {\small 0.93} & {\small 0.95} & {\small 0.96} & {\small 0.96} & {\small 0.97} \\
{\small\em Shot Noise} & {\small 0.51} & {\small 0.55} & {\small 0.58} & {\small 0.60} & {\small 0.62} \\
{\small\em Partial} & {\small 0.80} & {\small 0.80} & {\small 0.80} & {\small 0.72} & {\small 0.66} \\
{\small\em View} & {\small 0.63} & {\small 0.58} & {\small 0.60} & {\small 0.59} & {\small 0.59} \\
\hline
{\small{\textbf{Average}}} & {\small 0.64} & {\small 0.68} & {\small 0.70} & {\small 0.70} & {\small 0.71} \\
\hline
\end{tabular}
\caption{\small Quality of Scale invariant Spin Image feature description algorithm (average normalized $L_2$ distance between descriptors at corresponding points) on feature points detected using Mesh-Scale DoG (2). Average number of points: 1538.\label{tab:keller_spin-1.desc}}
\end{table}

%
%

\subsection{Dense feature descriptors}

Table~\ref{tab:Zobel_Valentin_GHKS} shows the performance of the GHKS dense feature description algorithm, in terms of normalized average $L_2$ distance between corresponding descriptors. 
Some results could not be computed by the participants.

\begin{table}
\centering
\begin{tabular}{lccccc}
& \multicolumn{5}{c}{\small\textbf{ Strength}} \\
\cline{2-6}
{\small\textbf{ Transform.}} & {\small\textbf{  1}} & {\small\textbf{ $\leq$2}} & {\small\textbf{ $\leq$3}} & {\small\textbf{ $\leq$4}} & {\small\textbf{ $\leq$5}}\\
\hline
{\small\em Isometry} & {\small 0.57} & {\small 0.58} & {\small 0.62} & {\small 0.62} & {\small 0.65} \\
{\small\em Rasterization} & {\small --} & {\small --} & {\small --} & {\small --} & {\small --} \\
{\small\em Sampling} & {\small 0.73} & {\small 0.74} & {\small 0.86} & {\small 0.94} & {\small 0.92} \\
{\small\em Holes} & {\small --} & {\small --} & {\small --} & {\small --} & {\small --} \\
{\small\em Micro holes} & {\small --} & {\small --} & {\small --} & {\small --} & {\small --} \\
{\small\em Scaling} & {\small 0.62} & {\small 0.61} & {\small 0.65} & {\small 0.65} & {\small 0.75} \\
{\small\em Affine} & {\small 1.08} & {\small 1.32} & {\small 1.46} & {\small 1.61} & {\small 1.77} \\
{\small\em Noise} & {\small 3.24} & {\small 3.37} & {\small 3.37} & {\small 3.34} & {\small 3.32} \\
{\small\em Shot Noise} & {\small 0.89} & {\small 1.03} & {\small 1.21} & {\small 1.33} & {\small 1.40} \\
{\small\em Partial} & {\small 0.80} & {\small 0.97} & {\small 1.12} & {\small 1.10} & {\small 1.15} \\
{\small\em View} & {\small --} & {\small --} & {\small --} & {\small --} & {\small --} \\
\hline
\end{tabular}
\caption{\small Quality of GHKS feature description algorithm (average normalized $L_2$ distance between descriptors at corresponding points).\label{tab:Zobel_Valentin_GHKS}}
\end{table}
%

%
%

\bibliographystyle{eg-alpha}

\bibliography{generic,desc/litman,desc/harris3D,desc/Zaharescu.bib,desc/smeets.bib,desc/zobel.bib,desc/Darom_Keller.bib}

\end{document}